\pdfoutput=1 

\documentclass[11pt,a4paper]{article}
\usepackage[acceptedWithA]{tacl2018v2}
\usepackage{nert}
\usepackage{times}
\usepackage{latexsym}
\usepackage{amssymb}

\usepackage{listings}
\usepackage{stmaryrd}
\usepackage{alltt}
\renewcommand{\UrlFont}{\ttfamily}

\usepackage{ccg-latex}

\usepackage{microtype}

\title{Supertagging the Long Tail \\with Tree-Structured Decoding of Complex Categories}  

\author{
Jakob Prange \quad Nathan Schneider \\
Georgetown University \\
\UrlFont{\{\emldisplay{jp1724@georgetown.edu}{jp1724}, \emldisplay{nathan.schneider@georgetown.edu}{nathan.schneider}\}@georgetown.edu} \And
Vivek Srikumar \\ 
University of Utah \\
\eml{svivek@cs.utah.edu}
}

\date{}

\begin{document}

\maketitle
\begin{abstract}
Although current CCG supertaggers achieve high accuracy on the standard WSJ test set, few systems make use of the categories' internal structure that will drive the syntactic derivation during parsing.
The tagset is traditionally truncated, discarding the many rare and complex category types in the long tail.
However, supertags are themselves trees.
Rather than give up on rare tags, we investigate constructive models that account for their internal structure, including novel methods for tree-structured prediction.
Our best tagger is capable of recovering a sizeable fraction of the long-tail supertags and even generates CCG categories that have never been seen in training, while approximating the prior state of the art in overall tag accuracy with fewer parameters.
We further investigate how well different approaches generalize to out-of-domain evaluation sets.
\end{abstract}

\section{Introduction}\label{sec:intro}

Combinatory Categorial Grammar \citep[CCG;][]{steedman-00} is a strongly-lexicalized grammar formalism
in which rich syntactic categories at the lexical level impose tight constraints on the constituents that can be formed.
Its syntax-semantics interface has been attractive for downstream tasks such as semantic parsing \citep{artzi-15} and machine translation \citep{nadejde-17}.

\begin{figure*}[t]
    \small
    \begin{subfigure}[t]{0.75\columnwidth}
      \centering\footnotesize\vspace*{-85pt}
      \cgex{5}{\hspace{7pt}\textcolor{violet}{Mary} & saw & \textcolor{teal}{John} & and & \textcolor{orange}{Bill}\\
        \cglines{5}\\
        \textcolor{violet}{\hspace{7pt}\texttt{NP}} & \texttt{(S\textcolor{violet}{\backs NP})\textcolor{olive}{\fs NP}} & \textcolor{teal}{\texttt{NP}} & \texttt{(\textcolor{olive}{NP}\textcolor{teal}{\backs NP})\textcolor{orange}{\fs NP}} & \textcolor{orange}{\texttt{NP}} \\
        }\\
      \caption{\label{fig:ccg-deriv} A CCG-supertagged sentence. Colors indicate functors and atomic categories that will unify in parsing.}
    \end{subfigure}\hfill
    \begin{subfigure}[t]{1.25\columnwidth}
      {\hfill
      \includegraphics[width=\textwidth]{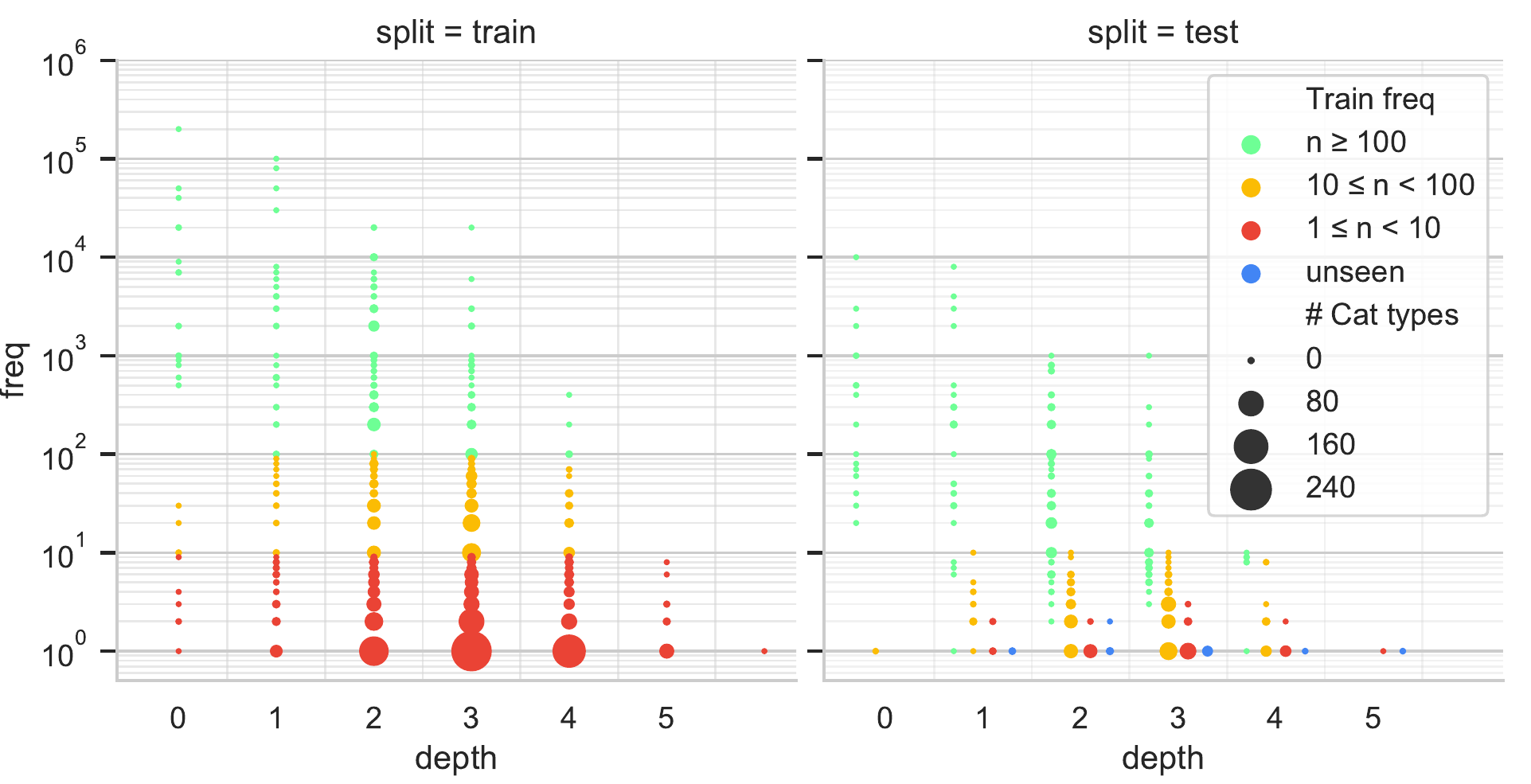}}
      \caption{\label{fig:types-tail}}
    \end{subfigure}
    \definecolor{shadecolor}{RGB}{255,255,255}
    \vspace{-15pt}
    \FrameSep0pt 
    \begin{shaded*}
    (b)\hspace{4pt} Number of supertag types (circle sizes) in relation to token log-frequency (y-axis) and supertag depth (x-axis) for the Rebank training set (left) and test set (right). Colors and horizontal offsets indicate supertags' training-data frequency band (decreasing frequency from left to right for each depth value).
    \end{shaded*}
    \vspace{-5pt}
    \caption{CCG supertags.}
\end{figure*}

Most CCG parsers operate as a pipeline whose first task is
`supertagging', i.e., sequence labeling with a large search space of complex `supertags' \citep[\emph{inter alia}]{clark-04,xu-15,vaswani-16}.
The complex categories specify valency information: expected arguments to the right are signaled with forward slashes, and expected arguments to the left with backward slashes.
For example, transitive verbs in English (like ``saw'' in \cref{fig:ccg-deriv}) are tagged \texttt{(S\backs NP)\fs NP} to indicate that they expect a subsequent object noun phrase (\texttt{NP}) and a preceding subject \texttt{NP} to form a clause (\texttt{S}).
Given the supertags, all that remains to parsing is applying general rules of (binary) combination between adjacent constituents until the entire input is covered.
Supertagging thus represents the crux of the overall parsing process.
In contrast to the simpler task of part-of-speech tagging, supertaggers are required to resolve most of the syntactic ambiguity in the input.

One key challenge of CCG supertagging is that the tagset is large and open-ended to account for combinatorial possibilities of syntactic constructions. 
This results in a heavy-tailed distribution of supertags, which is visualized in \cref{fig:types-tail}; a large proportion of unique supertags are rare or unseen (out-of-vocabulary, OOV) even in a training set as large as the Penn Treebank's.
Previous CCG supertaggers have surrendered in the face of this challenge: they treat categories as a fixed set of opaque labels, rather than modeling their compositional structure. 
Following \citet{clark-02}, the standard approach is to consider only supertags appearing at least 10 times in the training data,
sacrificing the possibility of predicting two thirds of the supertag types in CCGbank.
Rare supertags may have little impact on overall token accuracy---but the cost of this compromise is a fundamental incapability in truly generalizing to the task.

In this paper, we confront the long-tail problem head-on by proposing a \emph{constructive} framework in which supertags are built from scratch rather than predicted as opaque labels \citep{kogkalidis-19}. In contrast to prior  constructive supertaggers \citep{kogkalidis-19,bhargava-20}, our model builds upon the observation that supertags are themselves tree-structured, and hence can be generated top-down.\footnote{Our models and code are available at \url{https://github.com/jakpra/treeconstructive-supertagging}.}
Our experiments on the English CCGbank and its rebanked version show that constructing supertags as trees improves our ability to predict rare and even unseen tags, without sacrificing performance on the more common ones.

Our contributions are threefold:
\begin{enumerate}
    \item We introduce a general constructive supertagger that generates each lexical category recursively as a tree. To our knowledge, this is the first tree-structured predictor of its kind.
    \item We apply this model to English CCG supertagging. On frequent supertags, it matches the more traditional approach of using a fixed label set, while on the rare and unseen ones, we see substantial improvements in predictive performance.
    \item We perform an array of in-depth analyses that highlight the impact of different modeling and inference choices for the task of predicting supertags.
\end{enumerate}

\section{Motivation}\label{sec:background}

\begin{figure}[t]
    \centering\small\ttfamily
    \begin{tabular}{ll}
    Cat     & := FxnCat \textcolor{lightgray}{|} AtomCat \\
    FxnCat  & := Cat Slash Cat \\
    AtomCat & := N \textcolor{lightgray}{|} NP \textcolor{lightgray}{|} S \textcolor{lightgray}{|} PP \textcolor{lightgray}{|} ... \\
    Slash   & := \fs\ \textcolor{lightgray}{|} \backs \\
    \end{tabular}
    \caption{The `syntax' of CCG categories, using infix notation for complex categories (\texttt{FxnCat}). Our model generates supertags of type \texttt{Cat} top-down from this grammar.}
    \label{fig:cat-syn}
\end{figure}

\subsection{Anatomy of a Supertag}\label{sec:anatomy}

The internal structure of any CCG supertag is a tree licensed by the CFG in \cref{fig:cat-syn}.
Atomic categories like \texttt{S} and \texttt{NP} are related by slashes to form functional categories, which can in turn participate in larger functional categories.
By convention, the infix-notation supertag (\texttt{S}\textcolor{violet}{\backs \texttt{NP}})\textcolor{olive}{\fs \texttt{NP}} 
is equivalent to the tree in \cref{fig:cat-tree}, with prefix notation (\textcolor{olive}{\fs}~(\textcolor{violet}{\backs}~\texttt{S}~ \textcolor{violet}{\texttt{NP}})~ \textcolor{olive}{\texttt{NP}}),
where the slash signals the direction in which the category can combine, the right child of any slash is the argument, and the left child is the result of combining the category with its argument.
These hierarchical supertags constrain lexical item combination, e.g., specifying subcategorization of verbs for an object \textcolor{olive}{\texttt{NP}} to the right (\textcolor{olive}{\texttt{\fs}}).
This flexibility leads to infinite\footnote{But see \cref{sec:rel-work} for a discussion of how linguistic patterns limit the set of \textit{observed} tags.} possible supertags; 
in practice, they follow a power law distribution.
CCGbank (comprising the WSJ portion of the Penn Treebank) contains numerous rare supertags, including several that occur only in the test set.
Still others can be expected to occur in a much larger English corpus.

In previous work, CCG supertaggers have skirted this problem by ignoring the long tail of supertags: specifically, the ones occurring fewer than 10 times in the training set.
The consequences of such a threshold can be seen from
\cref{fig:types-tail}, which visualizes the distribution of supertag types in terms of depth (representing supertag complexity) and token frequency.
The supertags seen in training that would be ignored under a threshold of 10 appear in red, and the test set supertags never seen in training in dark blue. Though these only account for 0.2\% of tokens in the test set, they are present in nearly 4\% of sentences and represent fully two thirds of supertag \emph{types} in CCGbank.
Further, we see that rarer categories are increasingly more complex, i.e., their argument and result types are in turn composed of \texttt{FxnCats}.
Note in particular that the bulk of depth-4 categories and almost all categories with depth 5 or more fall below the 10-count threshold.

Inspired by the recent proposals of \citet{kogkalidis-19} and \citet{bhargava-20}, we hypothesize that modeling the structure of supertags, rather than treating them holistically and thresholding by frequency, can successfully generalize to rare and unseen tags.
For example, a good model should draw connections between words that are \texttt{NP}s themselves, words that take \texttt{NP}s as arguments (e.g., verbs), and words that yield \texttt{NP}s as their result (e.g., determiners).
We examine whether such linguistically-informed generalizations can benefit supertags of various frequency and structures, focusing on the rare and complex ones.

\subsection{Constructivity in Supertagging}\label{sec:constr-st}

We contrast two general paradigms for supertagging below. (Our experiments will explore multiple specific modeling strategies within each.)
 
\begin{figure}
    \centering\small
    \begin{subfigure}[t]{0.52\columnwidth}
      \centering\vspace{0pt}
      \includegraphics[width=\textwidth]{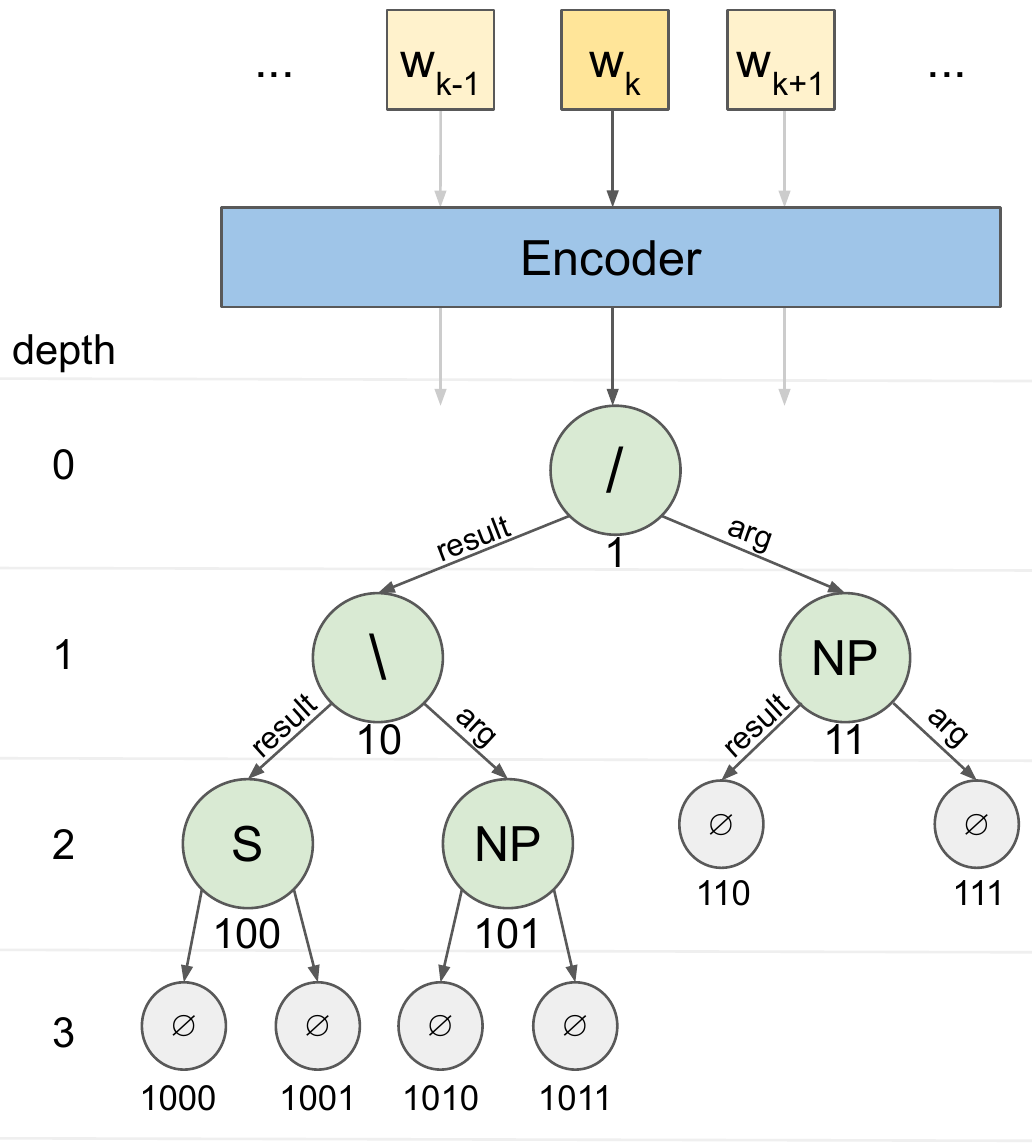}
      \caption{\label{fig:cat-tree}}
    \end{subfigure}
    \begin{subfigure}[t]{0.43\columnwidth}
      \centering\vspace{4pt}
      \begin{subfigure}[t]{\textwidth}
        \centering
        \includegraphics[width=0.8\textwidth]{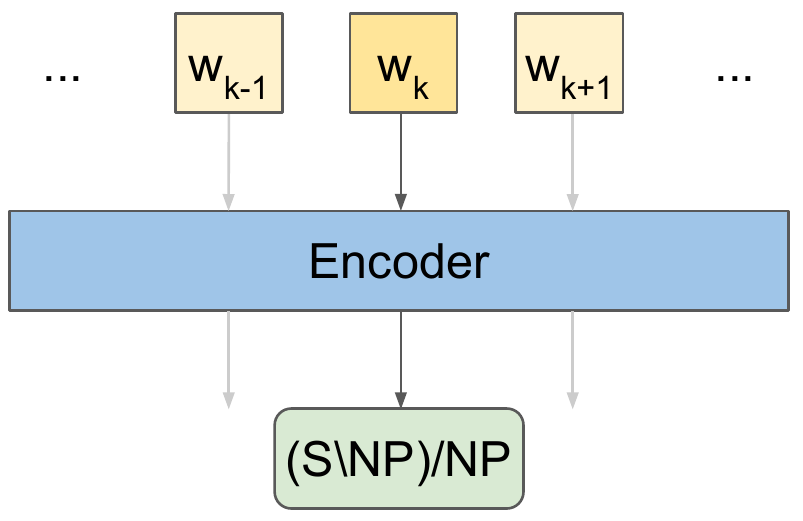}
        \caption{\label{fig:cat-bl}}
      \end{subfigure}\vspace*{14pt}
      \begin{subfigure}[t]{\textwidth}
        \centering
        \includegraphics[width=\textwidth]{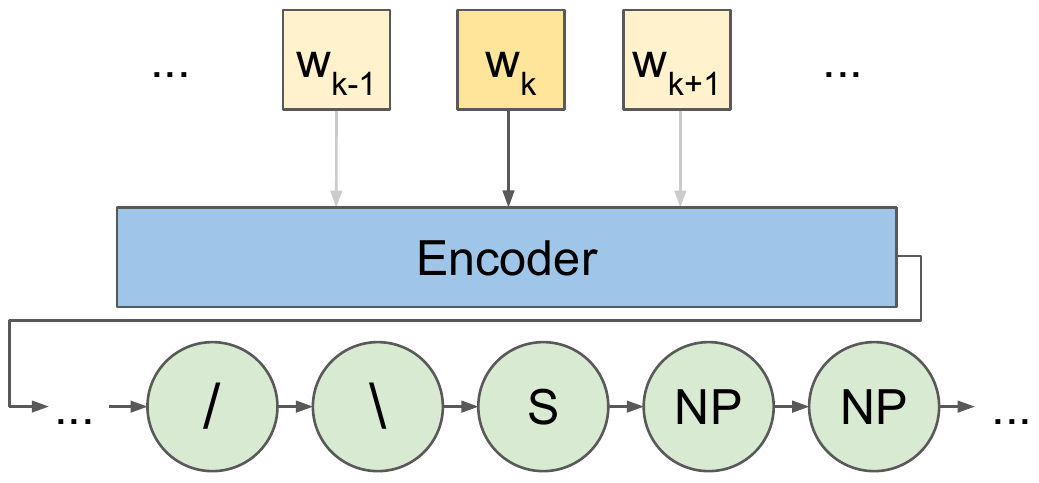}
        \caption{\label{fig:cat-seq}}
      \end{subfigure}
    \end{subfigure}\hfill%
    \caption{Schematic of our tree-structured supertagger (left) in contrast with unstructured (top right) and sequential (bottom right) models.
    Supertag depth also corresponds to decoding steps. Numbers below nodes denote positions or addresses.}
    \label{fig:cat-repr}
\end{figure}

Most previous supervised CCG supertaggers assume a closed tagset and \textbf{nonconstructively} assign one complete category per word (\cref{fig:cat-bl}).
This paradigm is oblivious to the internal structure of the supertag and incapable of predicting unseen supertags.
This is often combined with a frequency cutoff: 
only the $k$ supertags seen at least $n$ times in the training data 
are considered by the model, making each tag decision a $k$-way classification task.
Traditionally \citep{clark-02}, systems use a threshold of $n=10$ (yielding $k=425$ in CCGbank and $k=511$ in CCGrebank). The main motivation for this is to sidestep the most sparse and possibly noisy region of the output space without dramatically decreasing token coverage.
Below we experiment with both thresholded and non-thresholded models.

In contrast, a \textbf{constructive} tagger models the internal structure of supertags \citep{kogkalidis-19}. 
Supertags are constructed from minimal pieces (which for CCG are slashes and atomic categories).\footnote{For simplicity, we consider linguistic attributes like \texttt{dcl} (declarative) to be part of the atomic category.}
There is no frequency cutoff at training time.\footnote{In principle, a constructive model could be trained with frequency-thresholded training data, but we do not see any value in pursuing this option, as constructivity in itself already mitigates noise and sparsity.} %
At test time, supertags are predicted piece by piece, and there is no constraint that predicted supertags must have been seen before.
This can be done sequentially or recursively, taking the categories' internal tree structure into account.

Two different methods of sequential decoding have been explored by \citet{kogkalidis-19} (hereafter `K+19') and \citet{bhargava-20} (`BP20').
K+19 used a sequence-to-sequence model, with a single target sequence consisting of all serialized supertags for a sentence (\cref{fig:cat-seq}). They experimented with a type-logical grammar formalism similar to CCG, and a Dutch corpus. 
BP20 decoded CCG supertags as a separate sequence per token, and additionally conditioned each new supertag on the prediction history.
 
Here we go a step further and introduce methods for directly decoding supertags as \textit{trees}, freeing the models from having to learn this fundamental property from sequential data. We hypothesize that this will produce better and more compact representations that generalize to the long tail.

\section{Tree-Structured Constructive Supertagging}
\label{sec:method}

Given a sequence of words (a sentence), our goal is to predict each word's supertag.
Constructing a supertag from its components requires a scoring function for the parts that is cognizant of both surrounding words and categories.
Below we describe the decoding procedure (\cref{sec:inference}) and scoring functions (\cref{sec:model}) we developed for this purpose, which, in line with \cref{sec:background}, explicitly incorporate the categories' tree structure.

\subsection{Predicting Tree-structured Supertags}\label{sec:inference}

According to the grammar in \cref{fig:cat-syn}, each category is a binary tree with the following properties: 
\begin{inparaenum}[(1)]
\item  \texttt{Slash}es are non-terminals with two children: the category's argument (the syntactic type it seeks to combine with), and its result (the type it yields after combining with its argument). 
\item \texttt{AtomCat}s are leaf nodes. 
\item The root of the tree is either  the category's sole \texttt{AtomCat}, or its outermost functor, whose argument it seeks to combine with first.
\end{inparaenum}

Our output supertags are trees, but there is a crucial difference between our work and  constituency parsing of sentences.
In the latter case, the yield of a predicted tree is constrained to be the input sentence, thereby restricting both its depth and width. 
But in the case of supertagging, each word is associated with a binary tree--structured supertag whose breadth and depth are unknown at inference time.
We therefore grow supertags for each word from the top down (\cref{fig:cat-tree}).  
At the $t^{th}$ step, the model greedily chooses the most likely node labels at depth $t$, conditioned on the word encoding and the ancestors predicted so far (\cref{fig:cat-tree}). 
The first decision ($t=0$) is either an atomic category, or the main functor. In the latter case, the model then moves on to select the argument and result types, which may be atomic categories or functors themselves.
We are thus guaranteed to always generate well-formed categories.
As CCG supertags are not very deep in practice, we impose an upper limit on the depth of predicted trees based on the most complex categories found in the training and development data, with the main advantage that memory allocation during training can be bounded.\footnote{The limits on depth and arity are practical simplifications that follow from our task (supertags are always binary trees) and data distribution (there are no categories with depth $> 6$ in any of the training or development sets we use). However, our model can be generalized to trees of arbitrary depth, and not as easily, but conceivably, to a different or even variable arity. It turns out that none of the evaluation sets contain categories that are deeper that what is seen in training (except the redistributed test set in \cref{fig:shift-tail}, which contains one), so this measure has virtually no impact on tagging performance.}

\subsection{Modeling Supertags}\label{sec:model}

All supertagging models we compare consist of 
\begin{inparaenum}[(a)]
\item a sequence encoder, which generates a $d$-dimensional contextualized representation $\textbf{h}_{k, 0}$ for each word $k$ in a sentence $\textbf{x}$ (\cref{eq:enc}, together forming the $|\textbf{x}| \times d$ matrix $\textbf{H}_0$); %
\item an output-positional encoder, which generates the hidden representation $\textbf{h}_{k, i}$ for a position indexed by $i$ within the $k^{th}$ word's category tree; and %
\item a fully-connected 2-layer perceptron (MLP) with a final softmax layer which maps such a representation to a probability distribution $\textbf{o}_{k,i}$ over the inventory of possible labels $L$ (atomic categories and slashes; \cref{eq:out}). 
\end{inparaenum}
We use the term \textit{position} and the index $i$ to refer to any atomic part of a category for which a labeling decision has to be made. This could be, for example, the positions of the \texttt{S} category in \cref{fig:cat-seq,fig:cat-tree}, or the single output in \cref{fig:cat-bl}.
\begin{align}
\textbf{H}_{0} = & \operatorname{Encoder}\left(\textbf{x}\right) \label{eq:enc}\\
\textbf{o}_{k,i} = & \operatorname{MLP}\left(\textbf{h}_{k,i}\right) \label{eq:out}
\end{align}
The label $y_{k,i}$ is the most probable one per the MLP's prediction.

\paragraph{Contextualized word embeddings.}

In all conditions, we encode sentences using the pretrained RoBERTa-base encoder~\citep{liu-19-1},
finetuning it for our task.\footnote{We also experimented with a BiGRU encoder, but got consistently worse results.}
Several recent studies have shown that such models can capture syntactic properties and relations \citep[e.g.][]{jawahar-19,clark-19,hewitt-19}.

\paragraph{Output-positional encoding.}

We experiment with two alternative ways of deriving hidden states for category-internal positions $(k, i)$, where $i>0$:
a tree-structured recursive neural network \citep[\textit{TreeRNN};][\emph{inter alia}]{tai-15}, and a deterministic addressing function that accesses each node directly (\textit{AddrMLP}). 
Both variants, described below, also take into account the current node's ancestors.

The \textbf{TreeRNN} (\cref{eq:tree-rnn}) computes the hidden representation for a child node $c(i)$ from a vector embedding of its parent's label $y_{k,i}$ and the hidden representation $\textbf{h}_{k,i}$. The encodings are separately computed for child nodes representing the result ($c=\text{`left'}$) and argument ($c=\text{`right'}$) of the parent. Following K+19, we use the transpose of the last layer of the MLP to embed labels.
Our experiments use gated recurrent units \citep[GRUs;][]{cho-14}. 
\begin{align}
 \textbf{h}_{k,c(i)} = & \operatorname{GRU}_{c}\left(\operatorname{Embed}\left(y_{k,i}\right), \textbf{h}_{k,i}\right) \label{eq:tree-rnn}
\end{align}

Using tree-structured RNNs for top-down generation is reminiscent of \citet{zhang-16}.

For the \textbf{AddrMLP}, we represent the position $i$ of a node and the \texttt{Slash}es\footnote{Only \texttt{Slash} operators can have children (\cref{fig:cat-syn}).} in its ancestors (denoted by $\textbf{Y}_{k, \operatorname{anc}(i)}$) as a single feature vector that augments the contextualized word embedding:
\begin{align}
\textbf{h}_{k,i} = &  \textbf{h}_{k,0} + \operatorname{Linear}\left(\operatorname{Features}\left(i, \textbf{Y}_{k,\operatorname{anc}(i)}\right)\right) \label{eq:feat-rnn}
\end{align}

We employ a binary addressing scheme to refer to individual nodes: each node in a category's tree representation is addressed by a sequence of bits $a_0a_1a_2\dots a_T$, corresponding to a top-down traversal of the tree. The value $a_{t>0}=0\ (\text{or, } 1)$ is interpreted as branching to the left (or, right) at depth $t$. The root $a_0$ has an arbitrary placeholder value (say, $1$).\footnote{Prepending all addresses with 1 has several representational advantages, the most straightforward of which is that addresses can alternatively be read as binary numbers enumerating category pieces in breadth-first traversal.}
In the example in \cref{fig:cat-repr}, the inner \texttt{NP} argument (the argument of the top-level result) is addressed as $101$.
We represent the position of a node by a vector of elements in its address, mapping $a_{t>0}=0$ to $1$ and $a_{t>0}=1$ to $-1$ and ignoring $a_0$. The slashes in node's ancestors are similarly mapped to a vector consisting of $1$s for forward slashes and $-1$ for backward slashes. We use $0$ to pad feature vectors to a fixed maximum length.
We then use a single linear layer to project these features into the encoder's hidden space before adding it to the word's contextualized encoding.\footnote{The featurized encoder is, to a large extent, made possible by fixing the arity and maximum depth of categories. The TreeRNN will likely better admit more general setups, where outputs of unbounded depth and\slash or variable arity are allowed.}


\paragraph{Attention.}
While each word's contextualized encoding contains some information about all other words in the sentence, we hope to increase the model's output consistency using attention \citep{bahdanau-15,kim-17,wu-17} over the encoder's hidden state.
We compute attention weights $\alpha$ as in \cref{eq:attn-weights} and then add the $\alpha$-weighted context values to the hidden state, \cref{eq:attn-values}, replacing the simpler MLP from \cref{eq:out}.\footnote{We also tried self-attention over previously predicted partial outputs but did not find an increase in performance.}
\begin{align}
\alpha = & \operatorname{SoftMax}\left(\textbf{h}_{k, i}\textbf{H}_0^\top\right) \label{eq:attn-weights} \\
\textbf{o}_{k, i} = & \operatorname{MLP}\left(\textbf{h}_{k, i} + \alpha \textbf{H}_0\right) \label{eq:attn-values}
\end{align}

\subsection{Learning}\label{sec:learning}

We train the model using the AdamW optimizer \citep{loshchilov-19} and apply teacher forcing \citep{williams-89} to avoid a noisy feedback loop during learning.

\paragraph{Loss function.}

To achieve our goal of constructing correct and complete categories, we need our models to be correct in each atomic decision, even and especially for more complex categories.
We make the loss function sensitive to this by normalizing the cross-entropy between the predictions %
and the ground-truth %
only over the number of \textit{words} in a batch and retaining the unnormalized sum over individual atomic category decisions.
This naturally scales with category complexity.

If instead we were normalizing over atomic decisions, too, the loss contribution of, e.g., \texttt{NP} when it occurs inside a complex category \texttt{(S\backs NP)\fs NP} with size 5, would be 5 times smaller than when it occurs as a complete category on its own.
The disadvantage that complex categories already have as they tend to be rarer than simpler ones (\cref{fig:types-tail}) would be reinforced.
By keeping the atomic losses unnormalized, we therefore essentially put higher weight on the long tail in order to counterbalance this trend and improve generalizability.

\section{Experimental Setup}\label{sec:exp-setup}

Per our quest to \textit{supertag the long tail}, we compare our \textbf{TreeRNN} and \textbf{AddrMLP} models to the following baselines:
\begin{enumerate}[1)]
\item \textbf{Thresholded classification} (MLP\_10): We compute the output probabilities directly from the encoder's hidden state. (Since there is always exactly one output position for each input word, no additional encoding function is needed.) Only categories that are seen 10 times or more in training are considered. Supertags that fall below the threshold are replaced with an \texttt{<UNKNOWN>} symbol in training.
\item \textbf{Non-thresholded classification} (MLP\_1): Like MLP\_10, except that all tags seen in training may be predicted no matter their frequency.
\item \textbf{Per-sentence sequential} (K+19):
\citet{kogkalidis-19} construct type-logical supertags by generating for each sentence a single sequence of atomic types and functors (\cref{fig:cat-seq}). Trees are unwrapped in prefix notation and complete tags are separated from one another by a special token. %
We adapt K+19's implementation of the sequence-to-sequence Transformer model~\cite{vaswani-17}, accommodating its decoding procedure and memory requirements by training with a batch size of 32 for up to 256 epochs. We achieve the best performance using a cosine-annealed learning rate schedule that is warmed up over 10\% of the total training steps and with a warm restart after 128 epochs \citep{loshchilov-17}.
\item \textbf{Per-tag sequential} (RNN):
Instead of generating a single sequence for each sentence, \citet{bhargava-20} generate each word's supertag separately with an RNN. We implement a simplified version of \citeauthor{bhargava-20}'s model, omitting their prediction history connections between supertags, and using GRUs for decoding. We train this model for up to 50 epochs (batch sizes and learning rates are as with the tree-structured and nonconstructive models).
\end{enumerate}

If not indicated otherwise, we train the models with a batch size of 8 for a maximum of 10 epochs, and use early stopping based on the best development set performance.\footnote{Preliminary experiments showed that best dev performance is usually reached within 10 epochs; batches larger than 8 make our (single) GPU run out of memory.}
All reported results are averaged over 3 random restarts.

For downstream parsing evaluation (\cref{sec:analysis-parsing}),
we run the C\&C parser \citep{clark-07,clark-15} with the pretrained CCGbank model and default hyperparameters, providing as input our supertaggers' 1-best predictions and POS tags automatically obtained using Stanza \citep{qi-20}.

\begin{table}[t]
    \centering\small
    \begin{tabular}{lc|lcc}
Hidden dim $d$ & 768 & \multicolumn{2}{l}{Weight decay} & .01 \\
Activation & gelu & LRs & \multicolumn{2}{r}{1e-4, 1e-5 (ft)} \\
Dropout & .2 & Seeds & \multicolumn{2}{r}{14112, 36125,} \\
AdamW $\beta$'s & .9, .999 &  & \multicolumn{2}{r}{92225} \\
AdamW $\epsilon$ & 1e-6 & \multicolumn{2}{l}{Max cat depth} & 6 \\
    \end{tabular}
    \caption{Hyperparameters used in our experiments. We use separate learning rates (\textit{LR}) for fine-tuning (\textit{ft}) the RoBERTa-base model.}
    \label{tab:hyperparams}
\end{table}

\subsection{Model Details and Hyperparameters}\label{sec:hyperparams}

In \cref{tab:hyperparams} we report the model and training hyper\-para\-meters we use to facilitate replication of our results.
We performed  manual grid-search based on the development data to find workable learning rates. We chose a hidden dimensionality of 768 to match RoBERTa’s. We kept the default values for the AdamW hyperparameters.
We follow \citet{kogkalidis-19} in setting up the sequential Transformer model with 8 decoder heads and 2 decoder layers, but swap out the from-scratch encoder with RoBERTa-base.

\begin{table}[t]
    \centering\small
    \setlength{\tabcolsep}{4.4pt}
    \begin{tabular}{lrr}
         & \multicolumn{1}{c}{\textbf{CCGbank}} & \multicolumn{1}{c}{\textbf{Rebank}} \\ \midrule
cat types & 1,285 & 1,574 \\
\hspace{5pt} $\geq 100$ & 172 & 199 \\
\hspace{5pt} 10--99 (medium rare) & 253 & 312 \\
\hspace{5pt} $<$ 10 (very rare) & 860 & 1,063 \\
\hspace{5pt} atomic & 34 & 37 \\ \midrule
sentences & 39,604 & 39,604 \\
tokens & 929,552 & 943,204 \\
\hspace{5pt} medium rare cat & 7,549 & 9,640 \\
\hspace{5pt} very rare cat & 2,055 & 2,527 \\
    \end{tabular}
    \caption{Statistics of the CCG training corpora we use in our experiments.}
    \label{tab:data}
\end{table}

\subsection{Datasets}\label{sec:data}

We use two versions of the English CCGbank as in-domain (financial news) training and test sets: the original \citep{hockenmaier-07} and \citeposs{honnibal-10} `rebanked', i.e., corrected and enriched version (training sets reported in \cref{tab:data}; the results tables show test set counts).

The original CCGbank and Rebank differ in a number of conventions for atomic categories and category construction \citep{honnibal-10}. Rebank has a larger and more diverse category space, due in large part to a more principled treatment of NP argument structure.
Hence, we conduct our main experiments with Rebank and use the original CCGbank for comparisons with prior work.  

\begin{figure}[t]
    \centering
    \includegraphics[width=\columnwidth]{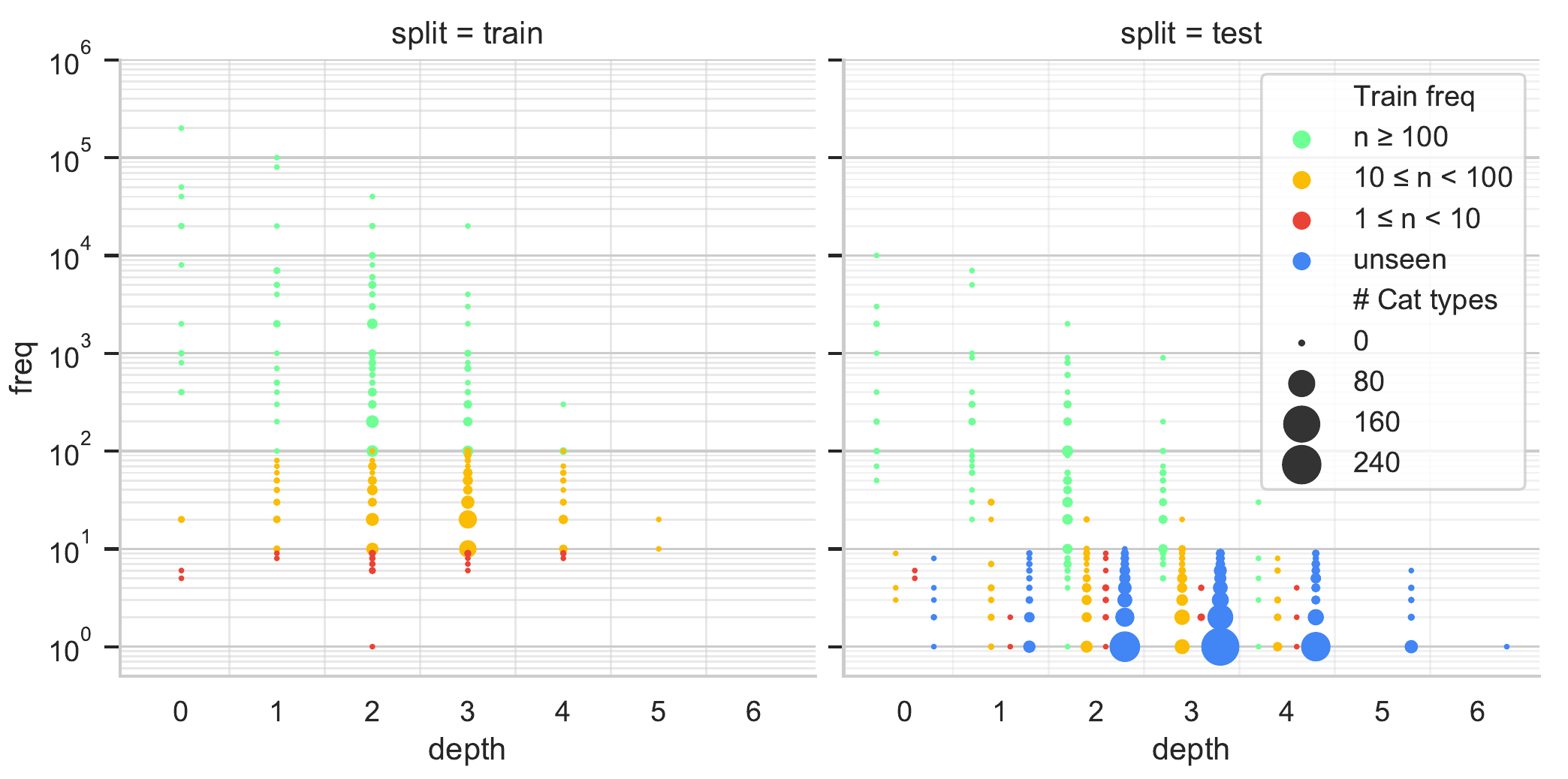}
    \caption{Shifting the tail to evaluation. The new test set (right) consists of those sentences in sections 02--21 that contain a category type occurring less than 10 times, and the new training set of the remaining sentences (left). As a result, we evaluate on many more category types that are not seen at all in training (dark blue circles\slash right-most horizontal offset for each depth) than before (\cref{fig:types-tail}).}
    \label{fig:shift-tail}
\end{figure}

A limitation of standard test sets for studying the long tail is that category types appearing rarely in training are even less frequent in evaluation (the Rebank test set contains just 107~tokens of categories seen 1--9 times in training, and only 27~tokens of OOV categories). Scores computed over these small samples may thus not reliably estimate the models' generalization capacity. 
We counteract this in two ways:
\begin{inparaenum}
\item by explicitly redistributing the training and test splits; and
\item by evaluating on out-of-domain data, with the assumption that a shift in domains means a shift in category distribution.
\end{inparaenum}

\begin{table*}[t]
    \centering\small
    \setlength{\tabcolsep}{3.2pt}
    \begin{tabular}{lc cccccccHHHH}
      & \textbf{Acc} & \multicolumn{4}{c}{\textbf{Acc by cat frequency in training 
        }} & \multicolumn{3}{c}{\textbf{Acc by cat depth}} & & & & \\ 
        \cmidrule(lr){2-2}\cmidrule(lr){3-6}\cmidrule(l){7-9} 
        & All       &  $\geq$100 & 10--99 & 1--9 & OOV & 0 & 1--2 & 3--6 & & & Parseability & LF \\  
        & \scriptsize $n$=56,395 & \scriptsize $n$=55,698    & \scriptsize $n$=563  & \scriptsize $n$=107    & \scriptsize $n$=27  & \scriptsize $n$=19,671 & \scriptsize $n$=33,409 & \scriptsize $n$=3,315 \\  
    \textbf{Model} & \scriptsize $N$=538 & \scriptsize $N$=199 & \scriptsize $N$=222 & \scriptsize $N$=91 & \scriptsize $N$=26 & \scriptsize $N$=18 & \scriptsize $N$=253 & \scriptsize $N$=267  \\ \midrule 
    \multicolumn{10}{@{}l}{\textbf{Nonconstructive Classification}} \\[3pt] 
    MLP\_10 & \textbf{94.77} \scriptsize\textcolor{gray}{$\pm$ .07} & \textbf{95.26} \scriptsize\textcolor{gray}{$\pm$ .07} & \textbf{68.32} \scriptsize\textcolor{gray}{$\pm$ 1.42} & -- & -- & \textbf{97.80} \scriptsize\textcolor{gray}{$\pm$ .14} & \textbf{94.25} \scriptsize\textcolor{gray}{$\pm$ .08} & 82.01 \scriptsize\textcolor{gray}{$\pm$ .18} & \textbf{94.81} \scriptsize\textcolor{gray}{$\pm$ .07} & 92.90 \scriptsize\textcolor{gray}{$\pm$ .12}  \\  
    MLP\_10@ & \textbf{94.76} \scriptsize\textcolor{gray}{$\pm$ .17} & \textbf{95.25} \scriptsize\textcolor{gray}{$\pm$ .18} & \textbf{68.98} \scriptsize\textcolor{gray}{$\pm$ 0.89} & -- & -- & \textbf{97.73} \scriptsize\textcolor{gray}{$\pm$ .16} & \textbf{94.29} \scriptsize\textcolor{gray}{$\pm$ .23} & 81.88 \scriptsize\textcolor{gray}{$\pm$ .29} & 94.80 \scriptsize\textcolor{gray}{$\pm$ .18} & 92.87 \scriptsize\textcolor{gray}{$\pm$ .46} \\[3pt]  
    MLP\_1   & \textbf{94.83} \scriptsize\textcolor{gray}{$\pm$ .09} & \textbf{95.27} \scriptsize\textcolor{gray}{$\pm$ .10} & \textbf{68.68} \scriptsize\textcolor{gray}{$\pm$ 1.09} & 23.99 \scriptsize\textcolor{gray}{$\pm$ 1.08} & -- & \textbf{97.71} \scriptsize\textcolor{gray}{$\pm$ .16} & \textbf{94.37} \scriptsize\textcolor{gray}{$\pm$ .14} & \textbf{82.39} \scriptsize\textcolor{gray}{$\pm$ .11} \\
    MLP\_1@ & \textbf{94.75} \scriptsize\textcolor{gray}{$\pm$ .18} & \textbf{95.18} \scriptsize\textcolor{gray}{$\pm$ .17} & \textbf{70.16} \scriptsize\textcolor{gray}{$\pm$ 0.81} & 27.10 \scriptsize\textcolor{gray}{$\pm$ 1.62} & -- & \textbf{97.84} \scriptsize\textcolor{gray}{$\pm$ .18} & \textbf{94.26} \scriptsize\textcolor{gray}{$\pm$ .52} & \textbf{82.33} \scriptsize\textcolor{gray}{$\pm$ .51} \\[3pt]
    \multicolumn{10}{@{}l}{\textbf{Constructive: Sequential}}  \\[3pt] 
    K+19 & 90.68 \scriptsize\textcolor{gray}{$\pm$ .15} & 91.10 \scriptsize\textcolor{gray}{$\pm$ .16} & 63.65 \scriptsize\textcolor{gray}{$\pm$ 0.21} & \textbf{34.58} \scriptsize\textcolor{gray}{$\pm$ 1.62} & \textbf{7.41} \scriptsize\textcolor{gray}{$\pm$ 0.00} & 91.71 \scriptsize\textcolor{gray}{$\pm$ .29} & 91.28 \scriptsize\textcolor{gray}{$\pm$ .02} & 78.43 \scriptsize\textcolor{gray}{$\pm$ .77} & 90.81 \scriptsize\textcolor{gray}{$\pm$ .16} & 84.97 \scriptsize\textcolor{gray}{$\pm$ .16} \\[3pt] 
    RNN & 93.92 \scriptsize\textcolor{gray}{$\pm$ .01} & 94.39 \scriptsize\textcolor{gray}{$\pm$ .02} & 65.48 \scriptsize\textcolor{gray}{$\pm$ 0.62} & 19.00 \scriptsize\textcolor{gray}{$\pm$ 2.35} & 0.00 \scriptsize\textcolor{gray}{$\pm$ 0.00} & 95.25 \scriptsize\textcolor{gray}{$\pm$ .09} & \textbf{94.33} \scriptsize\textcolor{gray}{$\pm$ .04} & \textbf{81.77} \scriptsize\textcolor{gray}{$\pm$ .78} \\
    RNN@ & 94.48 \scriptsize\textcolor{gray}{$\pm$ .08} & 94.93 \scriptsize\textcolor{gray}{$\pm$ .04} & 66.90 \scriptsize\textcolor{gray}{$\pm$ 2.32} & 27.41 \scriptsize\textcolor{gray}{$\pm$ 5.31} & 1.23 \scriptsize\textcolor{gray}{$\pm$ 2.14} & \textbf{97.72} \scriptsize\textcolor{gray}{$\pm$ .11} & 93.88 \scriptsize\textcolor{gray}{$\pm$ .09} & 81.33 \scriptsize\textcolor{gray}{$\pm$ .16} \\[3pt]
    \multicolumn{10}{@{}l}{\textbf{Constructive: Tree-structured}} \\[3pt] 
     TreeRNN & 94.62 \scriptsize\textcolor{gray}{$\pm$ .12} & \textbf{95.10} \scriptsize\textcolor{gray}{$\pm$ .11} & 64.24 \scriptsize\textcolor{gray}{$\pm$ 2.60} & 25.55 \scriptsize\textcolor{gray}{$\pm$ 0.54} & 2.47 \scriptsize\textcolor{gray}{$\pm$ 2.14} & \textbf{97.70} \scriptsize\textcolor{gray}{$\pm$ .21} & 94.14 \scriptsize\textcolor{gray}{$\pm$ .08} & 81.14 \scriptsize\textcolor{gray}{$\pm$ .90} & 94.66 \scriptsize\textcolor{gray}{$\pm$ .10} & \textbf{92.98} \scriptsize\textcolor{gray}{$\pm$ .84} \\  
     TreeRNN@ & 94.44 \scriptsize\textcolor{gray}{$\pm$ .22} & 94.95 \scriptsize\textcolor{gray}{$\pm$ .20} & 62.17 \scriptsize\textcolor{gray}{$\pm$ 3.03} & 22.43 \scriptsize\textcolor{gray}{$\pm$ 1.87} & 0.00 \scriptsize\textcolor{gray}{$\pm$ 0.00} & 97.61 \scriptsize\textcolor{gray}{$\pm$ .05} & \textbf{93.95} \scriptsize\textcolor{gray}{$\pm$ .33} & 80.61 \scriptsize\textcolor{gray}{$\pm$ .63} & 94.49 \scriptsize\textcolor{gray}{$\pm$ .22} & 92.50 \scriptsize\textcolor{gray}{$\pm$ .17}  \\[3pt]  
     AddrMLP & 94.58 \scriptsize\textcolor{gray}{$\pm$ .16} & 95.01 \scriptsize\textcolor{gray}{$\pm$ .16} & 67.44 \scriptsize\textcolor{gray}{$\pm$ 1.45} & \textbf{34.89} \scriptsize\textcolor{gray}{$\pm$ 2.35} & 3.70 \scriptsize\textcolor{gray}{$\pm$ 0.00} & \textbf{97.73} \scriptsize\textcolor{gray}{$\pm$ .13} & 94.02 \scriptsize\textcolor{gray}{$\pm$ .17} & 81.47 \scriptsize\textcolor{gray}{$\pm$ .24} & 94.62 \scriptsize\textcolor{gray}{$\pm$ .16} & 92.60 \scriptsize\textcolor{gray}{$\pm$ .35} \\  
     AddrMLP@ & \textbf{94.70} \scriptsize\textcolor{gray}{$\pm$ .05} & \textbf{95.11} \scriptsize\textcolor{gray}{$\pm$ .06} & \textbf{68.86} \scriptsize\textcolor{gray}{$\pm$ 0.57} & \textbf{36.76} \scriptsize\textcolor{gray}{$\pm$ 2.86} & 4.94 \scriptsize\textcolor{gray}{$\pm$ 2.14} & \textbf{97.85} \scriptsize\textcolor{gray}{$\pm$ .16} & 94.11 \scriptsize\textcolor{gray}{$\pm$ .03} & 81.92 \scriptsize\textcolor{gray}{$\pm$ .26} & 94.75 \scriptsize\textcolor{gray}{$\pm$ .05} & 92.18 \scriptsize\textcolor{gray}{$\pm$ .23} \\  
    \end{tabular}
    \caption{
    Main results on \textbf{Rebank} evaluation set (WSJ section 23).
    Accuracy scores are computed for bins based on the order of magnitude of category occurrences in training, and complexity of categories in depth, with depth=0 corresponding to atomic categories like \texttt{NP} (\cref{fig:cat-tree} has depth 2).
    Token ($n$) and type ($N$) counts for each bin are given in the first two rows.
    \textit{`@'} refers to model variants that use an attention mechanism over the encoder's hidden states.
    (As a Transformer model, the K+19 model attends to both the encoder and previously predicted outputs by default.)
    In each column, we \textbf{highlight} all results $r$ that fall within the standard deviation of the best result $b$, i.e., when $r + \operatorname{stdev}(r) > b - \operatorname{stdev}(b)$.
    For comparison, the overall tagging accuracy reported in \citet{honnibal-10} is {92.2\%}.
    }
    \label{tab:results-1}
\end{table*}

In the first case, we train the models on sentences containing exclusively the higher-frequency ($\geq$10) categories, and evaluate them only on sentences with at least one rare category. We split the usual Rebank training set (WSJ sections 02--21) in this way---the distribution follows \cref{fig:shift-tail}.\footnote{The few supertags in the 1--9 range of the new training set are those which occurred slightly above 9 times in the original training set, but some of their tokens were moved due to occurring in the same sentence as a low-frequency tag.}
In comparison with the default data splits (\cref{fig:types-tail}), we see that this sampling method captures precisely the long tail of categories, while leaving the rest of the category distribution largely unchanged.

For out-of-domain evaluation we use \citeposs{honnibal-09} (English) Wikipedia gold standard and the (English) {gold} section of the Parallel Meaning Bank, v3.0, which comprises multiple text types, including literary and biblical texts \citep[PMB;][]{abzianidze-17}.
The Wikipedia dataset follows CCGbank in terms of category conventions, while PMB is more similar to Rebank; we evaluate models trained on one style only on in- and out-of-domain test sets matching that style.
That said, PMB contains an unusually large number of unseen categories following idiosyncratic conventions that even Rebank-trained models are unlikely to pick up on without additional training data.

\section{Results}

We report our main results on Rebank in \cref{tab:results-1}.
In terms of overall accuracy, the tree-structured constructive supertaggers (best: 94.70\%) outperform the sequential ones (90.68\%, 93.92\%) and are roughly on par with the nonconstructive classifiers (best: 94.83).
Performance is generally very similar across all systems, except K+19. We conjecture that the main disparities between K+19 and the other models lie in the increased `cognitive load' of having to learn the correct structure of categories, as well as the missing hard alignment between words and supertags at test time.

\begin{table*}[ht]
    \centering\small
    \setlength{\tabcolsep}{4.4pt}
    \begin{tabular}{l ccccccc}
    \multicolumn{1}{c}{\textbf{}}   & \textbf{Acc} & \multicolumn{4}{c}{\textbf{Acc by cat freq in training 
        }} & \multicolumn{2}{c}{\textbf{Parsing}} \\
                         \cmidrule(lr){2-2}\cmidrule(lr){3-6}\cmidrule(l){7-8}
          & All       &  $\geq$100 & 10--99 & 1--9 & OOV & LF & Parseability  \\ 
                      & \scriptsize $n$=55,371  & \scriptsize $n$=54,825 & \scriptsize $n$=442 & \scriptsize $n$=82 & \scriptsize $n$=22 & & \scriptsize $n$=2,407 \\
        \textbf{Model} & \scriptsize $N$=435 & \scriptsize $N$=171 & \scriptsize $N$=176 & \scriptsize $N$=67 & \scriptsize $N$=21 \\
        \midrule 
        
    \multicolumn{8}{@{}l}{\textbf{Nonconstructive}} \\[3pt] 
    MLP\_10@ & {96.09} \scriptsize\textcolor{gray}{$\pm$ .07} & \textbf{96.50} \scriptsize\textcolor{gray}{$\pm$ .08} & \textbf{67.27} \scriptsize\textcolor{gray}{$\pm$ 1.02} & -- & -- & \textbf{90.78} \scriptsize\textcolor{gray}{$\pm$ .09} & 86.95 \scriptsize\textcolor{gray}{$\pm$ 0.75} \\
    MLP\_1 & \textbf{96.22} \scriptsize\textcolor{gray}{$\pm$ .06} & \textbf{96.58} \scriptsize\textcolor{gray}{$\pm$ .07} & \textbf{70.29} \scriptsize\textcolor{gray}{$\pm$ 2.35} & 23.17 \scriptsize\textcolor{gray}{$\pm$ 3.23} & -- & \textbf{90.91} \scriptsize\textcolor{gray}{$\pm$ .09} & 88.26 \scriptsize\textcolor{gray}{$\pm$ 0.39} \\[3pt] 
    \multicolumn{8}{@{}l}{\textbf{Constructive}}  \\[3pt] 
    K+19 & 92.12 \scriptsize\textcolor{gray}{$\pm$ .21} & 92.46 \scriptsize\textcolor{gray}{$\pm$ .20} & 65.38 \scriptsize\textcolor{gray}{$\pm$ 0.99} & \textbf{34.55} \scriptsize\textcolor{gray}{$\pm$ 4.28} & 1.52 \scriptsize\textcolor{gray}{$\pm$ 2.62} & 87.66 \scriptsize\textcolor{gray}{$\pm$ .19} & 91.14 \scriptsize\textcolor{gray}{$\pm$ 0.13} \\ 
    RNN@ & 95.10 \scriptsize\textcolor{gray}{$\pm$ .07} & 95.48 \scriptsize\textcolor{gray}{$\pm$ .07} & 65.76 \scriptsize\textcolor{gray}{$\pm$ 1.71} & 26.02 \scriptsize\textcolor{gray}{$\pm$ 0.70} & 0.00 \scriptsize\textcolor{gray}{$\pm$ 0.00} & 90.63  \scriptsize\textcolor{gray}{$\pm$ .04} & 89.53 \scriptsize\textcolor{gray}{$\pm$ 0.18} \\
    AddrMLP@ & \textbf{96.09} \scriptsize\textcolor{gray}{$\pm$ .07} & \textbf{96.44} \scriptsize\textcolor{gray}{$\pm$ .08} & \textbf{68.10} \scriptsize\textcolor{gray}{$\pm$ 1.38} & \textbf{37.40} \scriptsize\textcolor{gray}{$\pm$ 1.41} & \textbf{3.03} \scriptsize\textcolor{gray}{$\pm$ 2.62} & \textbf{90.79} \scriptsize\textcolor{gray}{$\pm$ .08} & 86.03 \scriptsize\textcolor{gray}{$\pm$ 1.72} \\
    \end{tabular}
    \caption{Results on the \textbf{original CCGbank} evaluation set (WSJ section 23). 
    The population $n$ for computing Parseability is the number of sentences in the test set.
    In each column, we \textbf{highlight} all results that fall within the standard deviation of the best result.
    }
    \label{tab:orig-results}
\end{table*}

\begin{table}[t]
    \centering\small
    \begin{tabular}{@{}l ccc cc}
                      & \multicolumn{3}{c}{\textbf{Acc}} & \multicolumn{2}{c}{\textbf{Parsing}} \\
                     \cmidrule(lr){2-4}\cmidrule(l){5-6}
    \textbf{Model} & All & $\geq 10$ & OOV & LF & P\slash ability  \\ \midrule
    \multicolumn{6}{@{}l}{\textbf{Nonconstructive}} \\[3pt] 
    V+16 & 94.24 & -- & -- & 88.32 & -- \\
    C+18 & 96.05 & -- & -- & -- & --  \\
    T+20 & -- & 96.39 & -- & 90.68 & -- \\[3pt]
    \multicolumn{6}{@{}l}{\textbf{Constructive}}  \\[3pt]  
    BP20 & 96.00 & -- & 5 & 90.9\phantom{0} & 96.2\phantom{0} \\ 
    \end{tabular}
    \caption{Relevant baselines reported in previous work (on the original CCGbank): \citet{vaswani-16}, \citet{clark-18}, \citet{tian-20}, and \citet{bhargava-20}.}
    \label{tab:baselines}
\end{table}

\begin{table}[t]
    \centering\small
    \setlength{\tabcolsep}{3.2pt}
    \begin{tabular}{@{}lcccccHHHH@{}}
         & \textbf{Acc} & \multicolumn{4}{c}{\textbf{Acc by cat freq 
        }} & \textbf{Prec} &  &  &  \\ 
        \cmidrule(lr){2-2}\cmidrule(l){3-6} 
         & All       &  $\geq$100 & 10--99 & 1--9 & OOV & Novel & 0 & 1--2 & 3--6 \\ 
         &  \scriptsize $n$=53,765 & \scriptsize $n$=50,754 & \scriptsize $n$=989  & \scriptsize $n$=292    & \scriptsize $n$=1,730 &  & \scriptsize $n$=18,370 & \scriptsize $n$=31,155 & \scriptsize $n$=4,240 \\
      \textbf{Model} & \scriptsize $N$=1,351 & \scriptsize $N$=188 & \scriptsize $N$=240 & \scriptsize $N$=118 & \scriptsize $N$=805 & & \scriptsize $N$=26 & \scriptsize $N$=511 & \scriptsize $N$=814  \\ \midrule

    \multicolumn{3}{@{}l}{\textbf{Nonconstructive}} \\[3pt]
    MLP\_10 & 88.76 & {92.86} & \textbf{55.71} & 13.24 & -- & -- & \textbf{97.02} & 88.48 & 55.04 \\ 
    MLP\_1  & 88.79 & \textbf{92.87} & 55.61   & 19.29 & -- & -- & \\[3pt]
    \multicolumn{3}{@{}l}{\textbf{Sequential}}  \\[3pt]
    K+19 & 80.20 & 83.49 & 47.72 & 25.11 & \textbf{11.62} &  \\
    {RNN} & 88.73 & 92.64 & 52.92 & 23.52 & \phantom{0}5.38 \\[3pt]
    \multicolumn{3}{@{}l}{\textbf{Tree-structured}}  \\[3pt]
    TreeRNN & 88.78 & 92.54 & 49.90 & 20.55 & \phantom{0}9.62 \\
    AddrMLP & \textbf{89.01} & 92.70 & 54.03 & \textbf{26.48} & 10.96 & 36.75 & 96.92 & \textbf{88.58} & \textbf{57.75} \\
    \end{tabular}
    \caption{Performance of the best systems (the variants \textit{with attention} for each paradigm) on redistributed Rebank train\slash test splits. Frequency bins are based on the new training set.}
    \label{tab:iv-oov-results}
\end{table}

\begin{table}[t]
    \centering\small
    \setlength{\tabcolsep}{1.6pt}
    \begin{tabular}{@{}l cHcccccH@{}}
                      & \multicolumn{1}{c}{\textbf{Wiki}} & & \multicolumn{6}{c}{\textbf{PMB}} \\
                     \cmidrule(lr){2-2}\cmidrule(l){4-9}
         & Acc & LF & All & $\geq$100 & 10--99 & 1--9 & OOV & Test  \\
                  &  \scriptsize $n$=4,151 &  & \scriptsize $n$=53,739 &  \scriptsize $n$=52,010 &  \scriptsize $n$=870 &  \scriptsize $n$=191 &  \scriptsize $n$=668 &  \scriptsize $n$=5,744  \\
    \textbf{Model} &  \scriptsize $N$=138 &  &  \scriptsize $N$=243 &  \scriptsize $N$=129 &  \scriptsize $N$=47 &  \scriptsize $N$=14 &  \scriptsize $N$=53 &  \scriptsize $N$=155  \\ \midrule
    \multicolumn{9}{@{}l}{\textbf{Nonconstructive}} \\[3pt] 
    MLP\_10 & \textbf{92.54} &  & 90.11 & \textbf{92.10} & 57.05 & -- & -- & \textbf{90.91} \\ 
    MLP\_1  & 92.31 & & \textbf{90.27} & \textbf{92.10} & \textbf{63.41} & 29.14 & -- &  \\[3pt]
    \multicolumn{9}{@{}l}{\textbf{Constructive}}  \\[3pt] 
    K+19 & 87.29 &  & 84.39 & 86.13 & 55.86 & 32.64 & 0.20 & 85.43 \\ 
    RNN & 92.00 &  & 89.52 & 91.38 & 61.42 & 24.26 & 0.25 & \\
    AddrMLP & 92.46 &  & {90.16} & 92.02 & {59.00} & \textbf{36.30} & \textbf{1.55} & 90.78 \\
    \end{tabular}
    \caption{Performance of the best systems (the variants \textit{with attention} for each paradigm) on the Wikipedia and PMB\footnotemark{} datasets. The state of the art on the Wikipedia data is 90.00\% \citep{xu-15}.}
    \label{tab:ood-results}
\end{table}

Regarding the long tail, we ask:
\emph{Can constructive models accurately predict rare and complex categories without sacrificing performance on the head of the distribution?}
To answer this question, we break down performance by the frequency of category types in the training data.
The baseline is the thresholded classifier MLP\_10, which performs well on frequent categories but cannot access rare categories occurring less than 10 times in training.
The simplest way of resolving this main hurdle is to remove the threshold, and indeed we find that MLP\_1 is able to predict about a quarter of long-tail categories correctly. Can we do better?
The sequence-to-sequence model by K+19 does a lot better on the tail and even retrieves some unseen categories, but at the cost of frequent ones. 
The per-tag recurrent and tree-recursive generators (RNN and TreeRNN) come close to to the nonconstructive classifiers, but do not convincingly improve over them. 
The AddrMLP model, finally, outperforms all others on the rare tail while matching nonconstructive taggers on frequent and simple ones.


\addtolength{\baselineskip}{0pt minus 1pt}

For comparison with existing work (\cref{tab:baselines}), we also report results on the original CCGbank (\cref{tab:orig-results}).
Our best constructive and nonconstructive models are on par with the previously reported state of the art in terms of overall accuracy.
\citet{tian-20} only report performance on categories seen at least 10 times in training, i.e., the union of our `$\geq 100$' and `10-99' bins; our top-3 results on this 
subset are MLP\_1:\ $96.37\%$, MLP\_10@:\ $96.27\%$, AddrMLP@:\ $96.22\%$.
The rise in absolute scores from \cref{tab:results-1} to \cref{tab:orig-results} is consistent with \citeposs{honnibal-10} finding that Rebank is more difficult to supertag and parse than CCGbank due to its sparser category space.
We therefore encourage future researchers to conduct experiments on Rebank and report detailed results for frequency- and complexity-binned subsets of the output space to facilitate more in-depth comparisons.

\addtolength{\baselineskip}{0pt minus-1pt}

\paragraph{Evaluating generalizability.}

One of the inherent problems of the supertagging task is the sparsity of the output space.
This is, however, not sufficiently captured by standard evaluation sets, as illustrated in \cref{fig:types-tail}.
To test how well the models \textit{really} generalize to the long tail, we evaluate them on alternatively sampled training and evaluation splits of the WSJ data (\cref{tab:iv-oov-results}) as  well as in domains diverging from the WSJ training set (\cref{tab:ood-results}).
These experiments largely confirm our findings from the standard Rebank evaluation set, while the change in category distribution has several important effects on our ability to evaluate model generalization:
First, OOV performance is much higher on the redistributed data (\cref{tab:iv-oov-results}) than on the standard test splits in \cref{tab:results-1,tab:orig-results,tab:ood-results}, highlighting all of the constructive models' generalization capability, and in turn suggesting that the OOV categories in WSJ section 23 and PMB are truly difficult, noisy, or otherwise inconsistent with the training data.
Second, the proportion of evaluation tokens of categories less than 10 times in training is 1.6\% in PMB and 3.8\% in our redistributed Rebank evaluation data, compared to only $\approx$0.2\% in the standard CCGbank and Rebank test sets. This 7x--16x increase in relative size renders the tail much more consequential for overall performance. And indeed we observe slightly smaller gaps in overall accuracy between the best-performing nonconstructive and the best-performing constructive systems in \cref{tab:ood-results} (0.08 on Wiki, 0.11 on PMB) compared to 0.13 in \cref{tab:results-1,tab:orig-results}, while in \cref{tab:iv-oov-results} AddrMLP even clearly outperforms the nonconstructive models.
Third, performance on rare and unseen categories can now be measured much more reliably due to the larger \textit{absolute} counts of rare and unseen categories. We provide in-depth analyses of this subset of tags in \cref{sec:analysis-unseen}.

In both in-domain and out-of-domain data, the performance gap between the nonconstructive MLPs and AddrMLP on the most frequent categories is minimal and in fact lies within the standard deviation.
Given the trend we observe from \cref{tab:results-1,tab:orig-results} to \cref{tab:ood-results,tab:iv-oov-results}, the ability to generalize to the long tail may well outweigh any minor improvement on the most frequent categories when applied to even more diverse data, within other languages, and across languages.

\footnotetext{Since we did not train any models on PMB itself, we analyze performance on all of PMB-gold, but for future comparisons, we also report accuracy on the suggested evaluation split: K+19:\ 85.43\%; RNN@:\ {90.24\%}; AddrMLP@:\ 90.78\%; MLP\_1@:\ {90.88\%}; MLP\_10@:\ {90.91\%}.}

\begin{figure}[t]
    \centering\small
    \includegraphics[width=0.87\columnwidth]{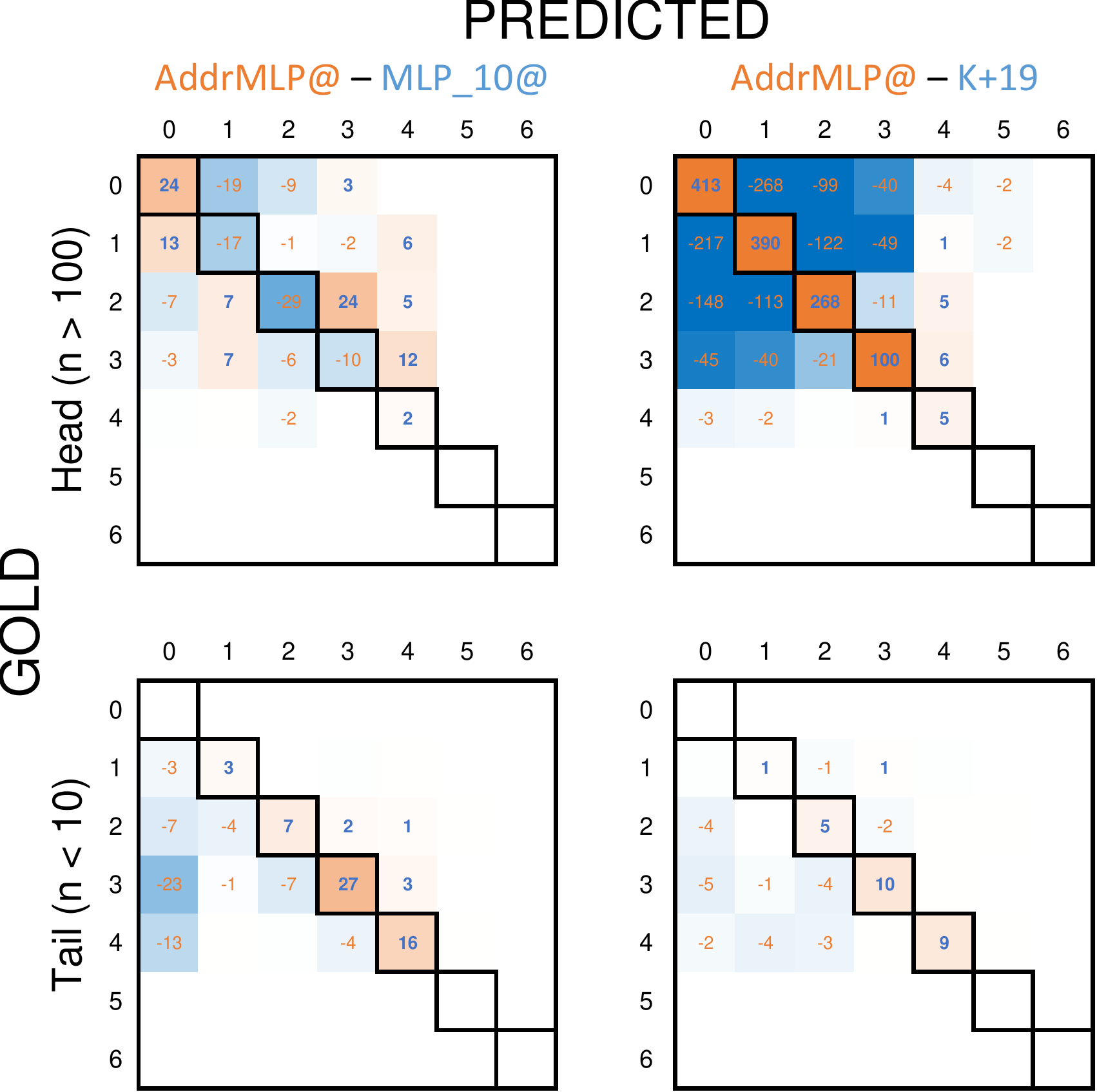}
    \caption{Confusion matrices by category depth, based on the standard Rebank evaluation set. Rows (columns) correspond to gold (predicted) categories with the respective depth. Thus, cells above (below) the diagonal refer to categories predicted too deep (shallow). All numbers are absolute differences between confusions made by AddrMLP@ and MLP\_10@ / K+19, respectively. Thus, positive numbers (red) are more typical for AddrMLP@ and negative numbers (blue) are more typical for one of the other systems.}
    \label{fig:depth-conf}
\end{figure}

\section{Detailed Analysis}\label{sec:analysis}

\subsection{Constructing Complex Categories}\label{sec:analysis-complex}

While nonconstructive taggers do not distinguish between categories of varying complexity (each supertag prediction is a single $k$-way decision), constructive taggers are always required to make multiple atomic decisions whenever assigning a complex category, all of which need to be correct in order for the full category to be counted as correct. This raises the question:
\emph{How difficult are categories of varying complexity for each of the systems?}

As \cref{fig:types-tail} shows, deeper, i.e., more complex categories tend to be rarer and thus are more difficult than simple ones in general, for all models.
Surprisingly however, we can see in the three rightmost columns of \cref{tab:results-1} that it is not dramatically more difficult for constructive systems to generate complex categories of depth $\geq 1$ than it is for nonconstructive systems to simply assign them (apart from K+19, which underperforms on frequent categories regardless of their complexity).

In \cref{fig:depth-conf} we take a closer look at the models' ability to predict categories of the appropriate depth. For the sake of brevity, we only consider three extreme cases: MLP\_10, K+19, and AddrMLP.
Compared with MLP\_10, which tends to choose one of the very frequent but relatively shallow categories of depth 1 or 2, AddrMLP prefers both standalone atomic (depth-0) categories and those of depth 3 and 4 (column totals in the top left matrix).
On the head, AddrMLP confuses depth-1 for depth-0 categories and overpredicts the depth of depth-2 and depth-3 categories more frequently than the baseline.
On the subset of rare categories (which are deeper than more frequent categories on average), AddrMLP is consistently better at predicting categories of the \textit{correct} depth (diagonal in the bottom left matrix); the thresholded model consistently chooses categories that are too shallow here.
The sequential tagger by K+19 struggles with predicting the correct depth for frequent categories much more than the tree-structured model (top right matrix), which is almost certainly a result of its lack of an inductive bias for the tree structure of categories. 
On the rare tail, however, its ability to guess the right depth is almost as good as that of AddrMLP (bottom right).

\begin{table}[p]
\centering
\footnotesize
\setlength{\tabcolsep}{2pt}
\begin{tabular}{|lccc|}
\hline
     & garnered & \multicolumn{1}{c}{from} & 1984 to 1986 \\ \hline 
    \textbf{Gold} & \textbf{\texttt{(S[pss]\textbackslash NP)}} & \textbf{\texttt{(ADV/ADV)/NP}} & \\
    MLP\_10 & \chk & \chk & \\
    MLP\_1 & \chk & \chk & \\
    K+19 & \chk & \chk & \\
    RNN & \chk & \chk & \\
    AddrMLP & \texttt{(S[pss]\textbackslash NP)/PP} & \texttt{(PP/ADV)/NP}  & \\
    \hline
\end{tabular}
\caption{AddrMLP treats ``garnered'' as expecting a PP argument (which would be correct for a source-PP, e.g.~``garnered information from the internet'', but this is a different sense of ``from'').
The other models correctly identify ``garnered'' as an intransitive passive verb with ``from'' introducing an adverbial PP adjunct. 
The gold category of ``from'' is so complicated because it is correlated with ``to'': First it expects an NP object on the right (``1984''), then an adverbial adjunct on the right (the to-PP), after which it produces an adjunct to a VP.\footnotemark{} AddrMLP's predictions for ``garnered'' and ``from'' are consistent in treating the entire construction ``from 1984 to 1986'' as an argument of the verb.}
\label{tab:ex-garnered}
\end{table}
\footnotetext{\texttt{ADV} is not an actual atomic category. We use it to abbreviate the VP-adjunct category \texttt{(S\textbackslash NP)\textbackslash (S\textbackslash NP)}. \texttt{PP} is a conventionalized atomic category for argument-PPs.}

\begin{table}[p]
\centering
\footnotesize
\setlength{\tabcolsep}{3.8pt}
\begin{tabular}{|lccc|}
\hline
      & orders began & piling & up \\\hline 
    \textbf{Gold} & & \textbf{\texttt{(S[ng]\textbackslash NP)/PR}} & \textbf{\texttt{PR}} \\
    MLP\_10 & & \texttt{S[ng]\textbackslash NP} & \texttt{ADV} \\
    MLP\_1 & & \texttt{S[ng]\textbackslash NP} & \texttt{ADV} \\
    K+19 & & \texttt{S[ng]\textbackslash NP} & \texttt{ADV} \\
    RNN & & \texttt{(S[ng]\textbackslash NP)/PP} & \texttt{S[adj]\textbackslash NP} \\
    AddrMLP & & \chk & \chk \\
    \hline
\end{tabular}
\caption{Here, the intended treatment of the particle (\texttt{PR}) ``up'' is as an argument selected by the predicate.
Only AddrMLP gets this right.
We assume this is preferable over treating it as a VP adjunct (as the nonconstructive and K+19 taggers do) from a semantic perspective, because ``pile up'' is a fixed expression with a meaning distinct from that of ``(to) pile'' or ``pile in''.
The RNN categories are both wrong and inconsistent (the ``piling'' category expects a PP and the ``up'' category is predicative).}
\label{tab:ex-piling}
\end{table}

\begin{table}[p]
\centering
\footnotesize
\begin{tabular}{|lccc|}
\hline
     & Why & constructive & ? \\\hline 
    \textbf{Gold} & \textbf{\texttt{S[wq]/(S[adj]\textbackslash NP)}} & \textbf{\texttt{S[adj]\textbackslash NP}} &  \\
    MLP\_10 & \texttt{<UNKNOWN>} & \chk & \\
    MLP\_1 & \texttt{(S/S)/(S[adj]\textbackslash NP)} & \chk &  \\
    K+19 & \chk & \chk &  \\
    RNN & \chk & \chk & \\
    AddrMLP & \chk & \chk & \\
    \hline
\end{tabular}
\caption{Supertags for \textsc{Wh}-words tend to be rare or unseen in training. Here, MLP\_10 correctly identifies that it cannot predict the true category for ``why'' and instead outputs \texttt{<UNKNOWN>}, while MLP\_1 chooses an incorrect tag. The constructive taggers are able to generate the correct category.}
\label{tab:ex-constructive}
\finalversion{\nss{footnote commenting on how this is an actual example (``But wielded by a pro like Jackie Mason, it is a constructive form of mischief. \textbf{Why constructive?} Because despite all the media prattle about comedy and politics not mixing, they are similar in one respect:'')}}
\end{table}

\subsection{Generation Behavior and Unseen Tags}\label{sec:analysis-unseen}

\emph{Are there any distinct patterns in the output of the different models?}
By manually searching the corpus, we find that even in the cases where a tagger assigns a category with an incorrect structure, there are systematic confusions such as between argument and adjunct PPs and between fixed particle verbs and (aspectual) adjunct particles. This is difficult to measure at a large scale, but we present two examples in \cref{tab:ex-garnered,tab:ex-piling}.

The thresholded tagger has the option to output an \texttt{<UNKNOWN>} label when it believes the correct category is not in the tagset.
It makes use of this option for 0.25\% of tokens on average (0.11\% with standard train\slash test splits); when it does, the correct category is indeed missing from the tagset about 2/3 of the time.  
This happens, e.g., with \textsc{Wh}-words in elliptical questions, as in \cref{tab:ex-constructive}.

\begin{figure}[t]
    \centering\small
    \begin{subfigure}[t]{.85\columnwidth}  
    \centering\vspace{0pt}
    \includegraphics[width=\textwidth]{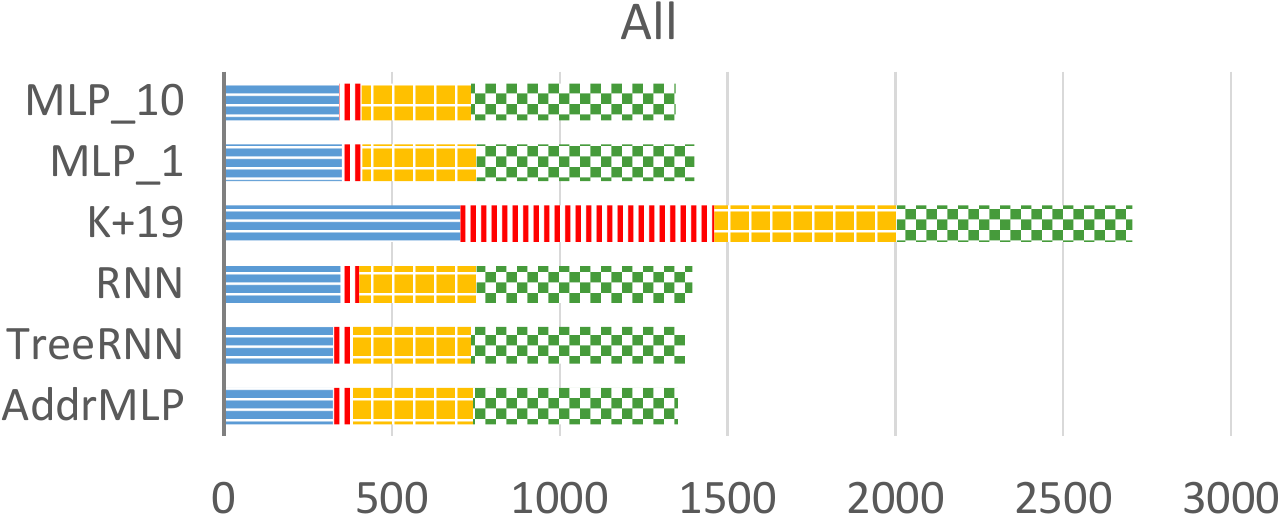}
    \end{subfigure}
    \begin{subfigure}[t]{.85\columnwidth}  
    \vspace*{7pt}
    \centering  
    \includegraphics[width=\textwidth]{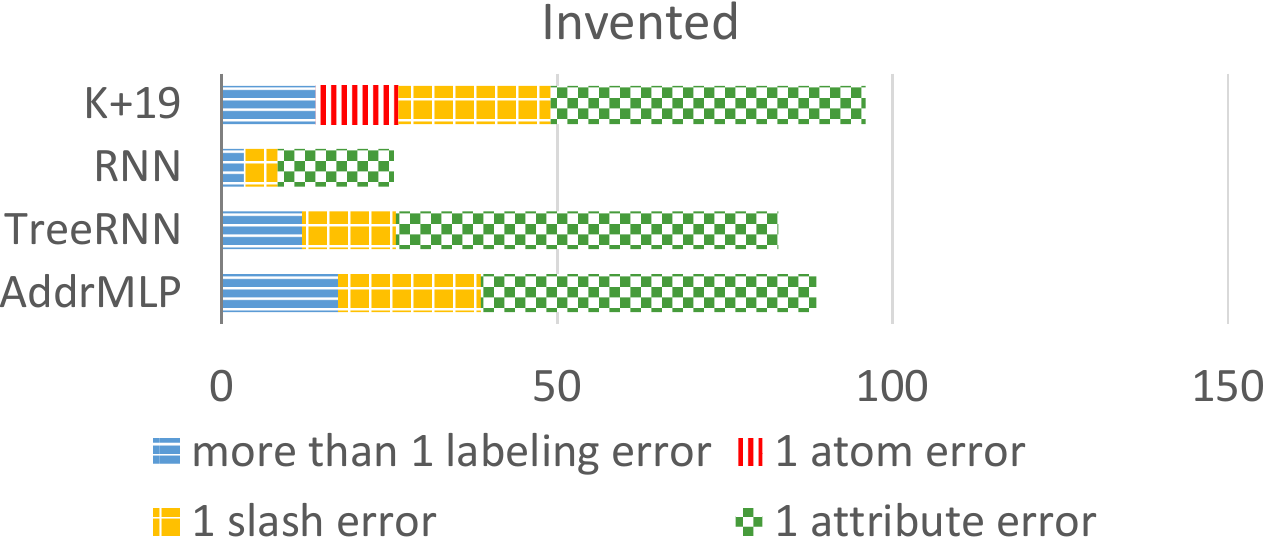}
    \vspace*{-10pt}
    \end{subfigure}
    \caption{Fine-grained analysis of correctly-structured but incorrectly labeled predictions (`\chk struct' in \cref{tab:structure-errs}).
    `Attribute error' means that the predicted atomic category is correct except for a wrong or missing linguistic attribute (e.g., \texttt{S} vs. \texttt{S[dcl]}); `atom error' means that an entirely wrong atomic category has been chosen (e.g., \texttt{PP} vs. \texttt{NP}); and `slash error' means confusing \texttt{\fs} and \texttt{\backs}.}
    \label{fig:fine-errors}
\end{figure}

In \cref{tab:structure-errs} we quantify the structural and labeling errors more generally, based on the redistributed evaluation set to ensure reliable estimates on rare phenomena.
A substantial portion of erroneous categories actually do have the correct \emph{structure} (\chk struct).\footnote{E.g., for ``piling'' in \cref{tab:ex-piling} the RNN predicts \texttt{(S[ng]\backs NP)\fs PP}, which exhibits the correct structure \texttt{(X\backs X)\fs X} with an incorrect atomic label (\texttt{PP} instead of \texttt{PR}).} For these cases, we perform a detailed error analysis, whose results we present in \cref{fig:fine-errors}.
In fact, if the structure is correct, the predicted category is often only off by the direction of a single slash or the attribute of a single atomic category.
K+19 additionally struggles with atomic decisions beyond just differences in attributes.

\begin{table}[t]
    \centering\small
    \setlength{\tabcolsep}{2.9pt}
    \begin{tabular}{clrrrrH}
         &      & \multicolumn{1}{c}{} & \multicolumn{3}{c}{\textbf{Incorrect}} \\
         \cmidrule(l){4-6}
         & \textbf{Model} & \multicolumn{1}{c}{\textbf{Correct}} & \multicolumn{1}{c}{\chk struct} & \multicolumn{1}{c}{\chk formed} & \multicolumn{1}{c}{\xxx formed}  \\ \midrule  
\parbox[t]{2mm}{\multirow{6}{*}{\rotatebox[origin=c]{90}{All}}} &  MLP\_10@ & 47,542 &     1,345 & 4,746 & -- & 133   \\
      &  MLP\_1@  & 47,552 &   1,401 & 4,811 & --  & --    \\
      &  K+19     & 43,120 &   2,706 & 7,812 & 127 & -- \\
      &  RNN@     & 47,704 &   1,395 & 4,661 &   5 & -- \\  
      &  TreeRNN@ & 47,733 &   1,373 & 4,659 &   1 & -- \\
      &  AddrMLP@ & 47,851 &   1,352 & 4,562 &   1 & -- \\ \midrule
\parbox[t]{2mm}{\multirow{4}{*}{\rotatebox[origin=c]{90}{Invented}}} &  K+19    & 201 & 96 & 160 & 127 & -- \\
      &  RNN@     &  93 &  26 &  71 &  5 & -- \\
      &  TreeRNN@ & 162 &  83 & 213 &  1 & -- \\
      &  AddrMLP@ & 190 &  89 & 240 &  1 & -- \\
    \end{tabular}
    \caption{Analysis of predicted supertag structures in the redistributed evaluation set. Incorrect predictions are broken down in terms of having the correct structure (\chk struct: the same number and arrangement of slashes, arguments, and results as the gold category), an incorrect but well-formed structure (\chk formed: diverging arrangement of arguments, but still obeying the grammar in \cref{fig:cat-syn}), or an invalid structure (\xxx formed, e.g., missing arguments to slashes).}
    \label{tab:structure-errs}
\end{table}

\emph{To what extent can the constructive models generate categories that were unseen during training?}
We take a closer look at categories the constructive taggers invented in the bottom halves of \cref{tab:structure-errs} and \cref{fig:fine-errors}.
K+19 is the most willing to invent categories, closely followed by the tree-structured models and finally RNN, which is rather conservative in this respect (see sums of the last four rows in \cref{tab:structure-errs}).
Merely generating more new categories irrespective of their correctness is of course not necessarily an advantage, but it is encouraging to see the models make use of their freedom to do so at an adequate rate, rather than only reproducing known categories or vastly overgenerating invented ones.
Interestingly, given that a incorrect invented category has the same structure as the gold category, we again see that the majority of errors are due to only a single attribute or slash, suggesting that in these cases the models get the general idea of the category right and only err in fine-grained and context-sensitive subcategorization. In the case of a slash mistake, they are notably also able to recover from it in later predictions.

\begin{table}[t]
\centering\footnotesize
\begin{tabular}{|lc|}
\hline
    & bring \\\hline 
    Predicted sequence & \texttt{/} \texttt{/} \texttt{\textbackslash} \texttt{S[b]} \texttt{NP} \texttt{\textbackslash} \texttt{NP} \\
    Predicted supertag & \texttt{((S[b]\textbackslash NP)/(NP\textbackslash \ \textbf{\_} ))/ \textbf{\_} } \\
    Gold sequence      & \texttt{/} \texttt{\textbackslash} \texttt{S[b]} \texttt{NP} \texttt{NP} \\
    Gold supertag      & \texttt{(S[b]\textbackslash NP)/NP} \\
\hline
\end{tabular}
\caption{A malformed supertag extracted from a sequence predicted by K+19. Underscores `\texttt{\_}' indicate gaps in the tree structure resulting from predicted surplus slashes.}
\label{tab:ex-malformed}
\end{table}

While the tree-structured taggers are guaranteed to produce valid categories,\footnote{That TreeRNN and AddrMLP still produced one malformed category can be considered a bug: They attempted to generate a category deeper than the maximally allowed depth and were unable to complete it. This is avoidable in practice.} it is possible for the sequential taggers to generate structurally invalid categories, i.e., sequences of atomic categories and slashes that are not licensed by the grammar in \cref{fig:cat-syn}.
With the tag-wise RNN generator, which generally refrains from inventing new categories, this only happens extremely rarely, but in the case of K+19, every 14th sentence is affected by an ill-formed supertag on average (every 66th sentence in the standard Rebank test set). 
A common source of errors is that too many slashes are predicted, whose argument and result slots can then not be filled by the predicted atomic categories.
We show an example in \cref{tab:ex-malformed}.



\begin{table}[t]
    \centering\small
    \setlength{\tabcolsep}{4pt}
    \begin{tabular}{l cc|c|c}                                   
    \cline{4-4}
          & \textbf{Nouns} & \textbf{Verbs} & \textbf{\textsc{Wh}} & \textbf{Other} \\
          & \scriptsize $n$=16,946 & \scriptsize $n$=7,915 & \scriptsize $n$=542 & \scriptsize $n$=29,968 \\
          & \scriptsize $N$=83 & \scriptsize $N$=296 & \scriptsize $N$=54 & \scriptsize $N$=436 \\
    \textbf{Model}  & \scriptsize $f$=1,158 & \scriptsize $f$=129 & \scriptsize $f$=38 & \scriptsize $f$=358 \\\midrule
    MLP\_10@ & 98.58 & 93.18 & 92.25 & 95.51 \\
    MLP\_1   & \textbf{98.62} & 93.49 & 92.68 & \textbf{95.65} \\
    K+19     & 95.58 & 90.54 & 90.04 & 90.62 \\
    RNN@     & 98.60 & 93.17 & 91.88 & 93.68 \\
    AddrMLP@ & 98.56 & \textbf{93.62} & \textbf{93.11} & 95.43 \\
    \cline{4-4}
    \end{tabular}
    \caption{Performance by part-of-speech, based on the original CCGbank test set.
    $n$ and $N$ refer to token and type counts in the test set, as before; $f$ refers to the average frequency with which a supertag belonging to the respective POS class is seen in training.}
    \label{tab:pos}
\end{table}


And vice versa, \emph{are there any categories that are not generated despite being seen in training?}
There are 80 category types in the standard Rebank test set that none of the tree-structured taggers ever predict correctly, although they are attested in the training data, and there are 93 types that are never retrieved by K+19, 73 of which overlap.
Out of these 73, no one occurs more than three times in the test set and almost all appear fewer than 50 times in training, with three exceptions: \texttt{(NP\backs NP)\backs (NP\backs NP)} (68 times in training), \texttt{((N\backs N)\backs (N\backs N))\fs NP} (50 times), and \texttt{(NP\backs NP)\fs N} (50 times). The first one is usually used for the last part of complex numerical expressions (such as dates and ranges), but the one token bearing this category in the test set is ``not'' in ``they might \textbf{not} miss one at all'', which is likely an annotation error.%
\footnote{There are a few more instances of such implausible lexical categories in the training data, like \texttt{S} or \texttt{((:\backs NP)\fs PP)\fs NP}.
}
The second one encodes prepositions modifying an appositive bare noun, typically an appellation or postposed proper noun.
The third one is for determiners of appositions or parentheticals.
67 of the 73 types that are problematic for the constructive models are never accurately predicted by the nonconstructive models either.


\subsection{Parts of Speech and Sentence Parsing}\label{sec:analysis-parsing}

Parsing performance is computed using labeled F1-score (LF) over CCG dependencies in all sentences, following \citet{clark-07}, and Parseability, i.e., the proportion of sentences for which a complete CCG derivation can be constructed.\footnote{The C\&C parser also reports coverage, the proportion of sentences for which at least one dependency relation can be recovered. Coverage is 100\% in all our conditions.}
Nearly all the models we compare outperform the state of the art in labeled dependency F1-score (right-most columns in \cref{tab:orig-results}). Interestingly, the K+19 model produces more parseable supertag sequences than others, despite consistently lagging behind in terms of category accuracy. Apparently this tagger prefers to be self-consistent over producing the actual correct categories, either due to its multihead attention mechanism, the fact that decisions towards the end of the sequence have access to all previously predicted categories in their entirety (rather than just parts of them), or both.

\begin{figure}[t]
    \centering
    \includegraphics[width=.97\columnwidth]{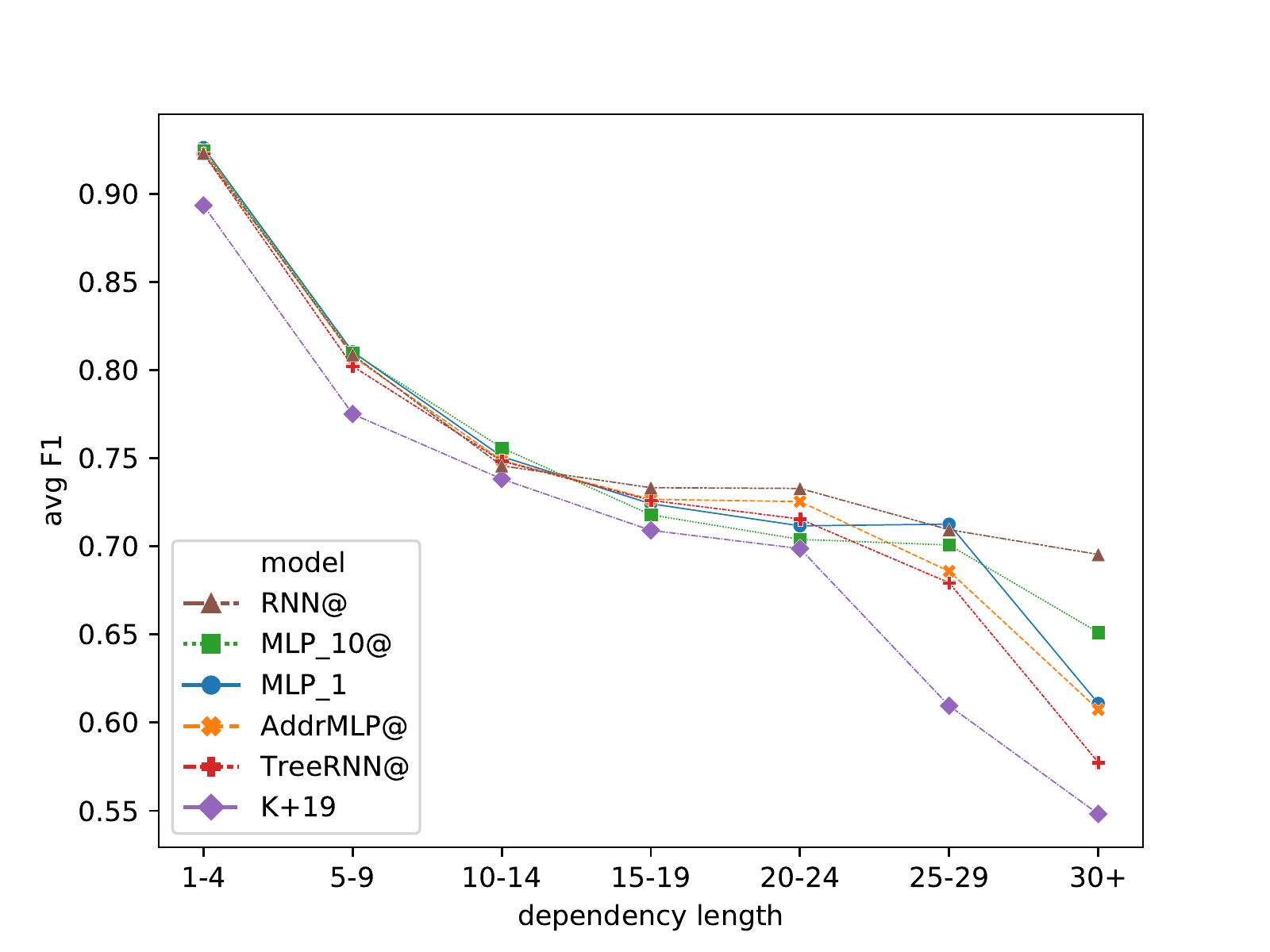}
    \caption{Parsing F1-score for varying dependency lengths, measured in terms of linear distance of the two words involved in the dependency.}
    \label{fig:dep-len}
\end{figure}

\paragraph{Long-range dependencies.}

We examine supertagging performance by POS class (a few are shown in \cref{tab:pos}) and find that constructive and nonconstructive taggers perform similarly across classes, with one notable exception: \textsc{Wh}-words, whose supertags are rarely seen in training and have a high type\slash token ratio at test time.
Their special syntactic status raises the question: 
\emph{How important are constructivity, tree structure, and long-tail recall for recovering categories involved in long-range dependencies?}

Somewhat surprisingly, we find that the RNN is best for these dependencies (\cref{fig:dep-len}), which might be related to the two parsing metrics in \cref{tab:orig-results}: RNN@ strikes a good balance between LF and Parseability. 
We further examine the average dependency length per category, and contrary to our expectation, dependencies involving \textsc{Wh}-categories are relatively short (usually 3--4 intervening words). 
We find that the supertags with the longest dependencies on average largely are functioning as subordinators, sentence adverbials, and inverted speech verbs such as \texttt{(S[dcl]\backs S[dcl])\backs NP}. 
These supertags have in common that they all contain sentential result\slash argument pairs of the form \texttt{S[$x$]|S[$x$]} (where $x$ is an optional attribute). The autoregressive nature of the RNN may be conducive to modeling the matching atomic categories of argument and result.
Exploring various decoding orders for both sequential and tree-structured constructive taggers in order to more explicitly take advantage of these intra-category relations is an interesting avenue for future work.
We also expect a major boost in Parseability from incorporating \textit{inter}-category prediction history into our models \citep{bhargava-20}. 
But this is nontrivial for tree-structured decoding and goes beyond our scope here.




\subsection{Runtime and Model Size}\label{sec:analysis-efficiency}

While the constructive taggers need to make more individual decisions for each supertag than nonconstructive ones, they only have to consider a much smaller and denser output space. This trade-off between time and space complexity should be considered in addition to tagging accuracy when evaluating each model.
Thus we ask: \emph{How do the constructive supertaggers compare to nonconstructive ones in terms of efficiency?}
In \cref{tab:analysis} we report model sizes (i.e., the number of learned parameters), training time until development performance plateaus, and inference speed.
As model size and runtime vary greatly between different constructive taggers, the answer to our question depends on how supertags are modeled and inferred.

\begin{table}[t]
    \centering\small
    \setlength{\tabcolsep}{4.2pt}
    \begin{tabular}{l ccc@{}}
      & \multicolumn{1}{c}{\textbf{Params}} & \multicolumn{1}{c}{\textbf{Train time}} & \multicolumn{1}{c@{}}{\textbf{Infer speed}} \\
                                    \cmidrule(lr){2-2}\cmidrule(lr){3-3}\cmidrule(l){4-4}
        \textbf{Model} &  \multicolumn{1}{c}{\textit{millions}} & \multicolumn{1}{c}{\textit{hours}} & \multicolumn{1}{c}{\textit{sents\slash s}}  \\ \midrule 
    \multicolumn{4}{@{}l}{\textbf{Nonconstructive Classification}} \\[3pt] 
    MLP\_10   & \hphantom{0}2.0 & \hphantom{00}9 & 191\hphantom{.0} \\  
    MLP\_1   & \hphantom{0}2.4 & \hphantom{0}11 & 195\hphantom{.0} \\[3pt]
    \multicolumn{4}{@{}l}{\textbf{Constructive: Sequential}}  \\[3pt] 
    K+19 & 11.8 & 120 & \hphantom{00}0.3 \\ 
    RNN   & \hphantom{0}4.8  & \hphantom{0}68  & 135\hphantom{.0} \\[3pt]
    \multicolumn{4}{@{}l}{\textbf{Constructive: Tree-structured}} \\[3pt] 
    TreeRNN   & \hphantom{0}8.3 & \hphantom{0}10 & 125\hphantom{.0} \\
     AddrMLP   & \hphantom{0}1.3 & \hphantom{0}10 & 126\hphantom{.0} \\
    \end{tabular}
    \caption{Model size and time required for training and inference. All models use the %
    RoBERTa encoder, whose 124.6 million parameters are not included here. Training times are approximate and include development set evaluation at every epoch.}
    \label{tab:analysis}
\end{table}

The K+19 sequential Transformer model has low efficiency for two reasons: The Transformer architecture itself has a large number of parameters; and sequential inference is slow because individual predictions for the same sentence cannot be parallelized and the number of inference steps per input sentence is linear in the sum of all category sizes (the number of atomic pieces) for that sentence.
The GRUs  in the RNN and TreeRNN models are much smaller than the Transformer of K+19, but the TreeRNN with its two GRUs for argument and result transitions ends up having almost as many parameters as the Transformer in total. 
The nonconstructive models map hidden representations into a much larger and sparser output space than the constructive models (and the output space of MLP\_1, in turn, is larger and sparser than that of MLP\_10).
AddrMLP, on the other hand, consists exclusively of feed-forward layers, resulting in the smallest model size among the ones we compare.

The sequential models require relatively many training epochs to converge. The reason total training time is still comparable between K+19 and RNN despite the extreme disparity in inference speed is that the Transformer is trained non-autoregressively and thus performs inference only \textit{between} epochs, for evaluation on the development set, whereas RNN training inherently relies on inference.
The nonconstructive and tree-structured models converge within the first 10 epochs.

For the per-tag constructive models RNN, TreeRNN, and AddrMLP we parallelize inference across all supertags in a batch, and for the tree-structured ones, we further parallelize the prediction of the children of slash functors, making their inference time logarithmic in the size of the largest predicted category in a sentence.

AddrMLP is both time- and space-efficient overall.
Its parameter count is only $\approx$1/10 of the K+19 model and $\approx$1/2 of the nonconstructive ones.

\section{Discussion \& Related Work}\label{sec:rel-work}

For a long time, researchers have addressed the large search space of CCG supertags.
\citet{baldridge-08} 
and \citet{ravi-10} 
were particularly concerned with high lexical ambiguity and counteracted this, respectively, by improving lexicon initialization using linguistic principles, and explicitly minimizing model sizes.
\citet{deoskar-13}, working with lexico-syntactic dependencies similar to supertags, addressed difficulties arising from the long tail of rare and unseen \textit{words}; and \citet{deoskar-14} addressed a similar issue specifically for generalizing a CCG parser.
The problem of out-of-vocabulary words has gotten much less severe with the advent of deep contextualized sentence encoders operating on subword units.

An alternative way of reducing the burden on the supertagger is to couple it with the parser and jointly optimize lexical and phrasal categories, subject to the combinatory rules of CCG \citep{auli-11,garrette-15}.
\citet{garrette-15} notably included a fully constructive probabilistic model of categories in a weakly-supervised grammar-induction scenario.
In the context of grammar induction for semantic parsing specifically, \citet{kwiatkowski-11} and \citet{artzi-15} have explored template-based methods to generalize a limited initial lexicon to likely alternative syntactic usages of observed words.

In the special case that all categories in a sentence but one are known, the combinatory rules of CCG can be reverse-engineered to infer the missing category. As an efficient and scalable example of this, \citet{thomforde-11} have proposed Chart Inference. 

Since the beginning of the neural era, virtually all advances in CCG supertagging have involved different means of deep sequence encoding, typically in the form of (Bi)LSTMs, with techniques including: predicting categories directly from the word-level encoder \citep{xu-15,lewis-16}; giving credit to likely category sequences \citep{vaswani-16,kadari-18}; forcing the model to distribute its attention over a fixed-size window of neighboring words \citep{wu-17}; training the encoder specifically to be aware of each word's neighboring categories \citep[`cross-view training';][]{clark-18}; and latently modeling parse chunks with a graph-convolutional network over word n-grams \citep{tian-20}.

Similar techniques have been applied to supertagging in the related formalism Tree-Adjoining Grammar (TAG) \citep{kasai-17,kasai-18}.
\citet{zhu-19} have formulated TAG supertagging as multitask learning with respect to certain aspects of the elementary trees' internal structure.
Their system predicts the category that optimizes the weighted sum of the scores for each subtask.

A possible objection to generating categories entirely productively is that universal linguistic patterns 
constrain the shape of categories and the syntactic relations they may engage in \citep{chomsky-93,baldridge-03}, and
%
for any given language, word order and other language-specific properties 
further restrict the underlying grammar.
Note that for a \texttt{FxnCat} shape with given argument and result types, the direction of its \texttt{Slash} functors is largely determined by global word order properties of the respective language.
Consider the prototypical category shape for adpositions, \texttt{(NP|NP)|NP}, where `\texttt{|}' stands for either forward or backward direction.
In English, a predominantly prepositional (as opposed to postpositional) language with postnominal-PP modifiers, this shape is most commonly instantiated as \texttt{(NP\backs NP)\fs NP}, but different ordering patterns may dominate in other languages.
Languages with more flexible word orders will show a greater variance in slash directionality than those with fixed word orders.
%
While our approach is in principle equipped to pick up on such patterns from data, we do not explicitly prohibit unlikely category types.
One potential way of incorporating such information is via logical constraints at training and\slash or inference time in the style of \citet{li-19,li-19-1}.
Another approach could be a hybrid one, bridging between constructive and nonconstructive tagging in a more fluid way.
We plan to explore these avenues in future work.

\section{Conclusion}

We introduced a novel, explicitly tree-structured CCG supertagging method, advancing the nascent paradigm of \textit{constructive} supertagging.
Our analysis of complex and long-tail categories highlights the positive impact of different modeling and inference choices within this paradigm: structural inductive bias as well as adequate contextualization via, e.g., attention contribute to more efficient, robust, and self-consistent models.
We hope that our proposed method can be instrumental in researching and applying not only CCG and related syntactic formalisms, but also other paradigms like morphological (de)composition of complex words in morphologically rich languages, or compositional semantic parsing.

\section*{Acknowledgements}

We would like to thank Aditya Bhargava and Konstantinos Kogkalidis for assistance with replicating their experiments and extended discussions of constructive models; Mark Steedman, Julia Hockenmaier, and Noah Smith for their deep insight into CCG; Kilian Evang and Lasha Abzianidze for explanations of data formats and conventions; Tao Li, Yichu Zhou, and Sean MacAvaney for help with implementing our models in PyTorch; and members of the NERT lab at Georgetown for feedback on an early abstract.
We are indebted to our TACL action editor Reut Tsarfaty and editor-in-chief Ani Nenkova, as well as the anonymous reviewers for their diligent assessment and handling of logistics.
This research was supported in part by NSF award IIS-1812778 and a generous gift from Google.





\bibliography{seq2tree}
\bibliographystyle{acl_natbib}

\end{document}